\documentclass[10pt,twocolumn,letterpaper]{article}

\usepackage{cvpr}              
\usepackage{amsmath,amssymb,amsthm}
\usepackage{graphicx}
\usepackage{booktabs}
\usepackage{multirow}
\usepackage{subcaption}
\usepackage{algorithmic}
\usepackage{nicefrac}
\usepackage{microtype}
\usepackage{xspace}
\usepackage{enumitem}
\usepackage{algorithm2e}
\usepackage{tabularx}
\newcommand{ \gainp}[1]{\textcolor{blue}{\mbox{\scalebox{0.7}{(#1)}}}}
\newcommand{ \gainm}[1]{\textcolor{red}{\mbox{\scalebox{0.7}{(#1)}}}}

\newcommand{ \gainu}[1]{\textcolor{blue}{\mbox{{#1}}}}

\usepackage{makecell}
\usepackage[table]{xcolor}
\usepackage{siunitx}

\definecolor{cvprblue}{rgb}{0.21,0.49,0.74}
\usepackage[pagebackref,breaklinks,colorlinks,allcolors=cvprblue]{hyperref}

\title{LARV: Data-Free Layer-wise Adaptive Rescaling Veneer for Model Merging}

\author{
Xinyu Wang\\
University of Connecticut\\
{\tt\small xinyu.wang@uconn.edu}
\and
Ke Deng\\
University of Georgia\\
{\tt\small ke.deng@uga.edu}
\and
Fei Dou\\
University of Georgia\\
{\tt\small fei.dou@uga.edu}
\and
Jinbo Bi\\
University of Connecticut\\
{\tt\small jinbo.bi@uconn.edu}
\and
Jin Lu\\
University of Georgia\\
{\tt\small jin.lu@uga.edu}
}

\begin{document}
\maketitle

\abstract{\normalfont
Model merging aims to combine multiple fine-tuned models into a single multi-task model without access to training data. 
Existing task-vector merging methods such as TIES, TSV-M, and Iso-C/CTS differ in their aggregation rules but treat all layers nearly uniformly. 
This assumption overlooks the strong layer-wise heterogeneity in large vision transformers, where shallow layers are sensitive to interference while deeper layers encode stable task-specific features. 
We introduce LARV, a training-free, data-free, merger-agnostic \textbf{L}ayer-wise \textbf{A}daptive \textbf{R}escaling \textbf{V}eneer that plugs into any task-vector merger and assigns a per-layer scale to each task vector before aggregation, and show it consistently boosts diverse merging rules.
LARV adaptively suppresses shallow-layer interference and amplifies deeper-layer alignment using a simple deterministic schedule, requiring no retraining or modification to existing mergers. 
To our knowledge, this is the first work to perform layer-aware scaling for task-vector merging.
LARV computes simple data-free layer proxies and turns them into scales through a lightweight rule; we study several instantiations within one framework (e.g., tiered two/three-level scaling with fixed values, or continuous mappings) and show that tiered choices offer the best robustness, while continuous mappings remain an ablation. LARV is orthogonal to the base merger and adds negligible cost. On FusionBench with Vision Transformers, LARV consistently improves all task-vector baselines across 8/14/20-task settings; for example, Iso-C + LARV reaches 85.9\% on ViT-B/32, 89.2\% on ViT-B/16, and 92.6\% on ViT-L/14. Layer-wise analysis and corruption tests further indicate that LARV suppresses shallow-layer interference while modestly amplifying deeper, task-stable features, turning model merging into a robust, layer-aware procedure rather than a uniform one.
}
\begin{figure*}[t]
    \centering
    \includegraphics[width=0.9\linewidth]{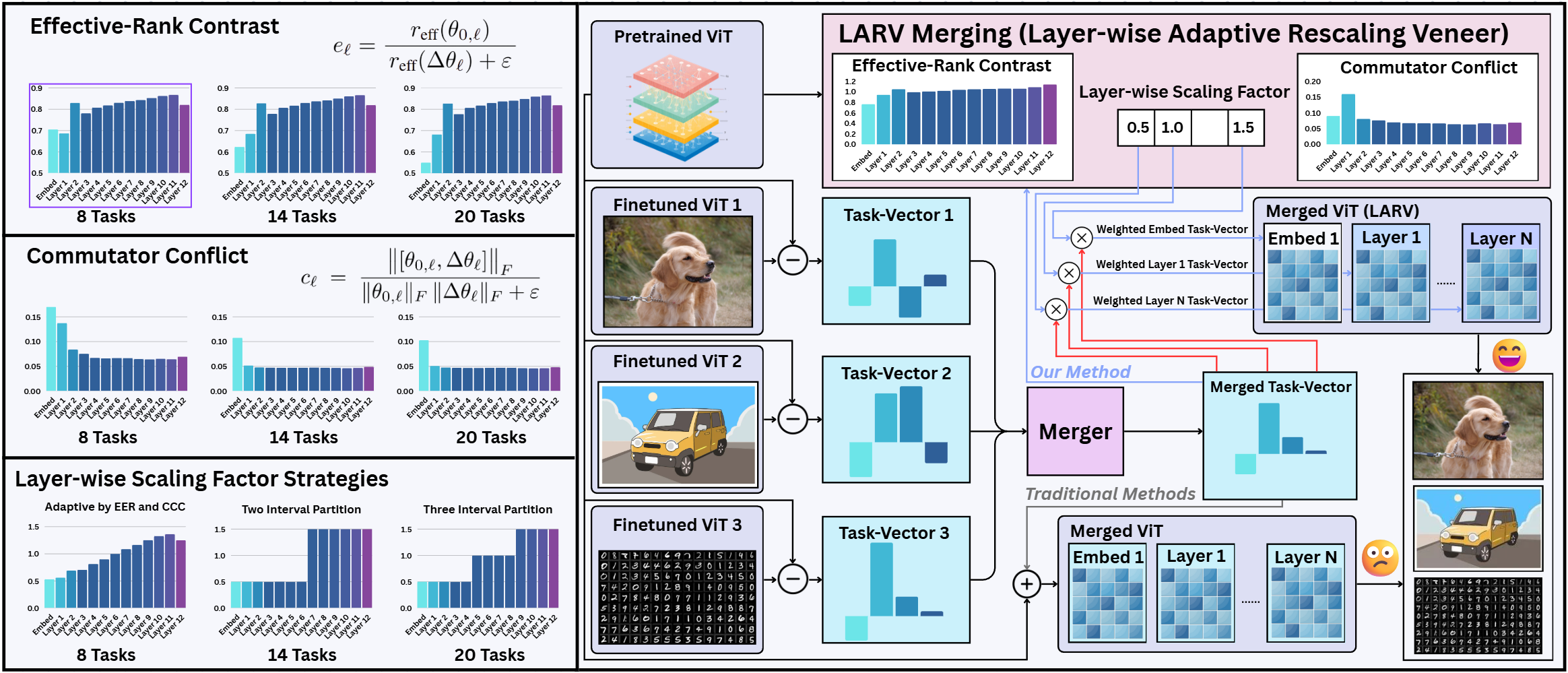}
    \caption{
Overview of LARV merging. (Left) We compute two data-free, weight-derived
diagnostics per layer—information richness $e_\ell$ and conflict level $c_\ell$—which
exhibit clear depth-dependent trends across 8/14/20-task merges. These signals
are converted into layer-wise scaling factors via continuous or tiered mappings.
(Right) Each task vector is rescaled at its corresponding layer and merged with
any base rule, forming a lightweight veneer that suppresses shallow-layer
interference while enhancing deeper-layer alignment.
}

\label{fig:lav_overview}
\end{figure*}
\section{Introduction}
\label{sec:intro}
\normalfont
Large pretrained vision models have made it practical to compose capabilities without retraining from scratch~\cite{bommasani2021opportunities,yurochkin2019bayesian,zheng2023preventing}. A convenient formalism is the task-vector view: a fine-tuned model for task $i$ at layer $\ell$ can be written as $\theta_{i,\ell}=\theta_{0,\ell}+\Delta\theta_{i,\ell}$, where $\theta_{0,\ell}$ is the pretrained initializer and $\Delta\theta_{i,\ell}$ is a task-specific delta. Model merging combines a set of task vectors $\{\Delta\theta_{i,\ell}\}$ into one vector $\Delta\theta_{\ell}$ and then adds it to $\theta_{0,\ell}$, so the merged model inherits multiple skills while preserving the base model’s generality. This paradigm is both meaningful for revealing surprising linear structure in parameter space and useful for enabling capability sharing with no data, rapid deployment under data-access constraints, and mix-and-match assembly across tasks and architectures~\cite{wortsman2022model,matena2022merging,ilharco2022editing,ainsworth2023gitrebasin,entezari2021lmc}. The ubiquity of weight averaging in practical distributed training further underscores its utility~\cite{mcmahan2017fedavg}.

Despite rapid progress, most merging strategies make \emph{global} choices that overlook depth~\cite{yang2024model}. Examples include a single mixing coefficient or checkpoint average~\cite{wortsman2022model}, model-level importance weighting via curvature/Fisher information~\cite{matena2022merging}, and uniform, rule-based schemes aimed at reducing interference (e.g., sign consensus, trimming, or parameter reweighting) that act similarly across all layers~\cite{yurochkin2019bayesian,yadav2023ties,du2024pcbmerging,ainsworth2023gitrebasin}. However, in deep networks this is a practical assumption: early layers predominantly encode high-frequency, local patterns that are information-dense yet fragile to noise and domain shift, whereas deeper layers tend to represent more semantic and stable features~\cite{yosinski2014transferable,zeiler2014visualizing,kornblith2019cka,raghu2017svcca}. This fragility at shallow depth has been repeatedly observed under common corruptions and shortcut biases~\cite{hendrycks2019benchmarking,geirhos2020shortcut}. When task vectors are combined without acknowledging this hierarchy, helpful updates in some layers can be weakened—or undone—by conflicting updates elsewhere, yielding unnecessary regressions~\cite{idelbayev2020low}.

This paper argues that merging should be layer-aware. Rather than treating every layer as equally trustworthy, we scale each layer’s contribution according to two properties: (i) how much \emph{information} the layer’s update appears to carry and (ii) how much \emph{conflict} it exhibits with other task vectors. Crucially, our solution is data-free: it requires no examples, gradients, or feature statistics and can be used wherever task vectors are available.

We introduce LARV (Layer-wise Adaptive Rescaling Veneer), a simple add-on that rescales the merged delta per layer before it is added back to the pretrained model. Given any base merger that produces $\Delta\theta_{\text{merge},\ell}$ for each $\ell\!\in\!\{1,\ldots,L\}$, LARV computes
\begin{equation}
\Delta\theta_{\text{larv},\ell} \;=\; r\!\left(e_{\ell},\,c_{\ell}\right)\cdot \Delta\theta_{\ell},
\end{equation}
where $r(\cdot)$ is a scalar gate determined by two \emph{data-free} diagnostics:
\begin{enumerate}
\item \textbf{Effective-Rank Contrast ($e_\ell$)}: a weight-only indicator of how
structured and informative the update is at layer $\ell$, with larger $e_\ell$
favoring concentrated, signal-bearing directions over diffuse noise.

\item \textbf{Commutator Conflict Coefficient ($c_\ell$)}: a measure of how
strongly the update interferes with the base operator, where larger $c_\ell$
indicates greater non-commutativity and potential conflict.

\end{enumerate}
Both quantities are computed directly from weights, with no data or gradients.
Empirically, $e_\ell$ increases and $c_\ell$ decreases with depth, motivating a
layer-wise schedule that down-weights fragile shallow layers and strengthens
stable deeper layers~\cite{yosinski2014transferable,zeiler2014visualizing}, yielding a lightweight veneer applicable to any
task-vector merger.

Our contributions are threefold:
\begin{enumerate}
\item LARV is \textbf{compatible with essentially all task-vector merging methods}, highlighting that structural (depth-aware) design matters and that layers should not share a single global scale.
\item We introduce novel indicators, $e_\ell$ (information richness) and $c_\ell$ (conflict level), to quantify per layer how much to encourage or discourage the merged update before recomposition, which yields a simple yet effective rule without acquiring data or gradients.
\item Across different base mergers and architectures, LARV yields consistent improvements as much as $7.7\%$ on FusionBench~\cite{tang2024fusionbench}, with detailed ablations, sensitivity analyses, and robustness evaluations to noise and distribution shift.
\end{enumerate}

\section{Related Work}
\label{sec:related}
\paragraph{Global and training-free merging.}
Early training-free methods merge parameters using \emph{global} rules that largely ignore depth.
Model Soups and weight averaging interpolate checkpoints without structural differentiation~\cite{wortsman2022model,izmailov2018averaging}.
Task Arithmetic treats the difference between a fine-tuned model and its pretrained ancestor as a task vector and composes vectors by addition or scaling~\cite{ilharco2022editing}.
Curvature-aware schemes such as Fisher-weighted averaging reweight parameters by importance but still apply a uniform rule across layers~\cite{matena2022merging}.
Regression Mean (RegMean) frames merging as a closed-form regression over layer inputs, again applying the same policy per layer~\cite{jin2022dataless}.
Permutation-alignment works (e.g., Git Re-Basin, permutation-invariance analyses) are complementary. They improve compatibility among independently trained models prior to averaging~\cite{ainsworth2023gitrebasin,entezari2021lmc}.

\paragraph{Structured and conflict-aware training-free merging.}
A second line adds structure to reduce interference while remaining training-free~\cite{sun2025cat,xiong2024multi}.
TIES-Merging trims small deltas and enforces sign consistency to avoid destructive averaging~\cite{yadav2023ties}.
DARE (Drop \& Rescale) sparsifies task vectors by randomly resetting most delta parameters to base values and rescaling the remainder before merging~\cite{yu2024language}.
EMR-Merging elects a unified task vector and attaches lightweight task-specific masks/rescalers applied at inference time~\cite{huang2024emr}.
TSV-M operates at the layer level, using singular vectors of per-layer task matrices to compress and whiten interference across tasks~\cite{gargiulo2025task}.
These methods demonstrate the utility of structural priors for dataless merging, yet most still apply a near-global policy to each layer’s update scale.

\paragraph{Adaptive or data-involving merging.}
Other approaches \emph{learn} coefficients with auxiliary signals~\cite{ye2023merging}.
AdaMerging optimizes task-wise or layer-wise coefficients by entropy minimization on unlabeled test data, improving generalization under shift~\cite{yang2023adamerging}.
Twin-Merging learns routers to dynamically integrate shared and exclusive expertise across tasks~\cite{lu2024twin}.
Some recent conflict-aware designs (e.g., CAT-Merging) remain training-free but rely on parameter-type and devise specific trimming and projections rather than learned gates~\cite{sun2025cat}.
While these techniques increase flexibility, methods that learn coefficients typically require samples' features and labels or considerable amount of extra compute at merge time.

\paragraph{Layer-wise representation studies and our position.}
A large body of work characterizes the \emph{hierarchical} organization of deep networks:
lower layers emphasize local/high-frequency patterns, while deeper layers encode more semantic and stable features~\cite{zeiler2014visualizing,yosinski2014transferable,raghu2017svcca,kornblith2019cka}.
Fragility at shallow depth under corruptions and shortcuts is well documented~\cite{hendrycks2019benchmarking,geirhos2020shortcut}.
Despite this, most mergers set a single global scale per model or per parameter type.
LARV occupies the middle ground:
it is \emph{training-free} like global methods, yet explicitly layer-aware via two data-free diagnostics—$e_\ell$ (information richness) and $c_\ell$ (conflict level) that deterministically gate each layer’s update before recomposition.
This yields interpretability and negligible overhead while capturing depth heterogeneity that global rules overlook.

\section{Method}
\label{sec:method}

\subsection{Preliminaries and Problem Setup}
\label{sec:prelim}
We consider a pretrained network $\theta_0$ with $L$ parametrized layers $\{\theta_{0,\ell}\}_{\ell=1}^{L}$. 
Each task $i\!\in\!\{1,\dots,K\}$ provides a \emph{task vector} (delta) $\Delta\theta_{i} \triangleq \theta_i-\theta_0$ anchored at $\theta_0$, with layer-wise slices $\{\Delta\theta_{i,\ell}\}_{\ell=1}^{L}$. 
Task vectors may be full-rank updates or PEFT modules (adapters, LoRA, BitFit), which we fold into the ambient parameter space for merging~\cite{chen2022adaptformer,houlsby2019adapters,hu2021lora,ben-zaken2022bitfit,zhao2024merging}. 
A base merger $\mathcal{M}$ aggregates the per-layer deltas
\begin{equation}
\Delta\theta_{\ell} \;=\; \mathcal{M}\big(\{\Delta\theta_{i,\ell}\}_{i=1}^{K}\big), \qquad \ell=1,\dots,L,
\label{eq:base-merge}
\end{equation}
where $\mathcal{M}$ may be arithmetic/averaging (Model Soups, SWA)~\cite{wortsman2022model}, curvature-aware (Fisher-weighted)~\cite{matena2022merging}, sign/trim-consistent (e.g., TIES-Merging)~\cite{yadav2023ties,qi2025less}, regression-style ~\cite{jin2022dataless,nguyen2025regmeanpp}, or permutation-aware alignment (Git Re-Basin, permutation invariance)~\cite{ainsworth2023gitrebasin,entezari2021lmc}. 
Classical training-free merging then injects $\Delta\theta_{\ell}$ using a \emph{single global} coefficient $s$, as shown below in Eq.~(\ref{eq:uniform}), implicitly assuming that all layers are equally trustworthy. Formally, we recall the standard uniform composition used throughout the merging literature:
\begin{equation}
    \theta_{\text{merge},\ell} = \theta_{0,\ell} \,+\, s \cdot \Delta \theta_{\ell},
    \label{eq:uniform}
\end{equation}
where $s\!\in\![0,\infty)$ controls the global blend (e.g., $s{=}1/K$ yields an even average for merging $K$ models). 

Eq.~\eqref{eq:uniform} treats every layer’s update equally, which is a convenient  assumption; however, a global coefficient can be fragile~\cite{wang2024rethinking}. A large body of empirical analysis shows that deep networks organize features hierarchically: earlier layers emphasize local/high-frequency patterns while later layers encode more semantic, task-relevant structure~\cite{zeiler2014visualizing,yosinski2014transferable,raghu2017svcca,kornblith2019cka}. Consequently, a single scalar $s$ \textit{over-updates fragile early layers and under-updates decisive later layers}, even when $\mathcal{M}$ is improved by curvature or sign heuristics. 
Permutation-aware methods~\cite{ainsworth2023gitrebasin,entezari2021lmc} mitigate \emph{representational misalignment} but the application of the update remains uniform across depth. This motivates the layer-wise veneer we introduce next.

\paragraph{Why a single global scale is insufficient.}
Consider a minimal two-layer network $f(x)=W_2\phi(W_1x)$. Suppose two tasks $A/B$ produce updates such that $\Delta W_1^{A} \approx -\Delta W_1^{B}$ (strong early-layer conflict) while $\Delta W_2^{A} \approx \Delta W_2^{B}$ (head agreement). Any global coefficient $s$ scales both layers identically and therefore cannot suppress the conflicting layer-1 update while preserving the aligned layer-2 update; the optimum requires $s_1 \neq s_2$. This simple counterexample illustrates that uniform scaling cannot resolve depth-localized interference, motivating the need for layer-wise scaling.

\subsection{LARV: Layer-wise Rescaling Veneer over Any Merger}
\label{sec:larv_design}
\paragraph{Design goals.}
Given the insufficient treatment of depth-wise heterogeneity across depth, we seek a veneer that is \emph{training-free} and \emph{label-free}, (ii) is an merger-agnostic add-on that can wrap any base rule $\mathcal{M}$, and (iii) explicitly encodes \emph{depth heterogeneity} consistent with representation studies~\cite{zeiler2014visualizing,yosinski2014transferable,raghu2017svcca,kornblith2019cka}.

Specifically, given the merged update $\Delta\theta_\ell$ at layer $\ell$ from Eq.~\eqref{eq:base-merge}, LARV's goal is to compose a \emph{layer-specific} scale:
\begin{equation}
    \theta_{\text{merge},\ell}
    \;=\;
    \theta_{0,\ell} \;+\; s_\ell \cdot \Delta\theta_\ell,
    \qquad \ell=1,\ldots,L,
\label{eq:lav}
\end{equation}
where $s_\ell\!\in\!\mathbb{R}_{\ge 0}$ is chosen without any training or labels.

From weights alone, we compute two per-layer diagnostics:
(i) an \emph{information-richness} proxy $e_\ell$ (Effective-Rank Contrast ; see \S\ref{sec:weight-only}) and
(ii) a \emph{conflict} proxy $c_\ell$ (Commutator Conflict Coefficient).
Then, based on the two diagnostics, a \emph{composite weight-only score} is formulated by encouraging high information and penalizing conflict in the following subsection.

\subsubsection{Weight-Only Metrics}
\label{sec:weight-only}

\noindent\textbf{Setup.} As in \S\ref{sec:prelim}, each parameterized module is viewed as a matrix: convolutions are folded into a $({\rm out}\times{\rm in}\cdot k_hk_w)$ matrix; attention uses $Q/K/V/O$ projections as separate parameter groups for metric computation and averages their scores to one per-layer value. Biases and LayerNorm scale/shift are included in the final composition (Eq.~\eqref{eq:lav}) but excluded from the spectral/commutator diagnostics to avoid degenerate shapes. We denote the base weights by $\theta_{0,\ell}$ and the merged update by $\Delta\theta_\ell$ (Eq.~\eqref{eq:base-merge}).

\paragraph{(A) Effective-Rank Contrast ($e_\ell$).}
Spectral entropy separates structured, low-rank updates from diffuse, noise-like ones. 
Let $\{\sigma_j(A)\}$ be the singular values of a matrix $A$; define normalized energies 
$p_j(A)=\sigma_j(A)^2/\|A\|_F^2$ and spectral entropy $H(A)=-\sum_j p_j(A)\log p_j(A)$. 
The \emph{effective rank} is $r_{\text{eff}}(A)=\exp(H(A))$, which is scale-invariant and larger for spectrally diffuse matrices. 
We compare the update to the base via a base-relative contrast:
\begin{equation}
e_\ell
=
\frac{r_{\text{eff}}(\theta_{0,\ell})}{\,r_{\text{eff}}(\Delta\theta_\ell)+\varepsilon\,},
\label{eq:eer}
\end{equation}
with a small $\varepsilon>0$ for numerical stability.
Thus $e_\ell$ is larger when $\Delta\theta_\ell$ is \emph{more concentrated} (lower effective rank) than the base $\theta_{0,\ell}$, which is interpreted as a more informative, task-salient update, and it shrinks toward $0$ when $\Delta\theta_\ell$ is as or more diffuse than the base.

We estimate $r_{\text{eff}}(\cdot)$ per layer using a lightweight randomized SVD (rank $k\in[32,64]$), applied to each layer’s matrix view. 
This preserves the training-free nature and adds negligible overhead relative to a forward pass. 
(If $\|\Delta\theta_\ell\|_F$ is numerically zero, we set $e_\ell=0$.)

This aligns with findings that deeper layers exhibit more semantic, low-intrinsic-dimension structure~\cite{raghu2017svcca,kornblith2019cka}.  As shown in the first column of Fig. \ref{fig:weight-metrics}, $e_\ell$ typically \emph{increases} with depth on ViT backbones (see the first column in Fig.~\ref{fig:weight-metrics}). 

\paragraph{(B) Commutator Conflict Coefficient ($c_\ell$).}  Order sensitivity (non-commutativity) between the base operator and the update acts like a rotation of features and signals potential interference. We quantify it by a scale-normalized commutator:
\begin{equation}
c_\ell
\;=\;
\frac{\big\|[\theta_{0,\ell},\Delta\theta_\ell]\big\|_F}
     {\|\theta_{0,\ell}\|_F\,\|\Delta\theta_\ell\|_F+\varepsilon},
\label{eq:ccc}
\end{equation}
with a small $\varepsilon\!>\!0$ for numerical stability and we define the operator between two matrices as
$[A,B]=AB-BA$. When $\theta_{0,\ell}$ and $\Delta\theta_\ell$ are rectangular, we use the average of left/right Gram-commutators (dimension-compatible and symmetric):
{\small
\begin{equation}
c_\ell \;=\; \frac{1}{2}\frac{\big\|\theta_{0,\ell}\Delta\theta_\ell^{\!\top}-\Delta\theta_\ell\theta_{0,\ell}^{\!\top}\big\|_F}{\|\theta_{0,\ell}\|_F\,\|\Delta\theta_\ell\|_F+\varepsilon}
\;+\; \frac{1}{2}\frac{\big\|\theta_{0,\ell}^{\!\top}\Delta\theta_\ell-\Delta\theta_\ell^{\!\top}\theta_{0,\ell}\big\|_F}{\|\theta_{0,\ell}\|_F\,\|\Delta\theta_\ell\|_F+\varepsilon}.
\label{eq:ccc-rect}
\end{equation}
}
This measures non-commutativity of the base operator and the update: if they commute, composition order doesn’t matter; if not, the update “rotates” features, indicating conflict. Large $c_\ell$ indicates destabilizing rotations and calls for shrinkage of the model update.

This quantity serves conceptually to measure commutativity and we will use it directly to penalize non-commutativity between $\theta_{0,\ell}$ and $\Delta \theta_\ell$. Empirically, the conflict $c_\ell$ tends to be higher in shallow layers and significantly decreases with depth across almost all base mergers, as one can observe in the second column in Fig. \ref{fig:weight-metrics}.


\subsubsection{Composite weight-only score and gate}
\label{sec:composite-gate}

We synthesize information (large $e_\ell$), conflict (large $c_\ell$), and a simple depth prior ($r_\ell$) into a single weight-only score defined in Eq.~\eqref{eq:composite}:
{\small
\begin{equation}
  w_\ell
  \;=\;
  \frac{\,e_\ell^{\,\eta}\; r_\ell^{\,\rho}\,}
       {\big(\mathrm{sp}(z(c_\ell))+1\big)},
  z(x_\ell)=\frac{x_\ell-\mu_x}{\sigma_x},
  \mathrm{sp}(t)=\log(1+e^{t}),
  \label{eq:composite}
\end{equation}
}
where $z(\cdot)$ standardizes a scalar across layers with mean $\mu_x$ and standard deviation $\sigma_x$, and $\mathrm{sp}(\cdot)$ is softplus. A multiplicative “product-of-experts” form lets $e_\ell$ and $c_\ell$ act jointly while $r_\ell$ encodes the empirical depth trend. We standardize $c_\ell$ across layers via $z(\cdot)$ so its scale is comparable model-to-model, pass it through softplus to keep the penalty positive and smooth, and add $+1$ to avoid vanishing denominators while making the mapping monotone in $c_\ell$. $\eta,\rho,\zeta\!\ge\!0$ tune the relative influence of each factor. In the paper we choose $\eta{=}1,\rho{=}0.5$.

After obtaining $w_\ell$, we map it to the per-layer scale using either a robust \emph{tiered} rule (default) or a continuous mapping.

\begin{figure}[t]
    \centering
    \includegraphics[width=\linewidth]{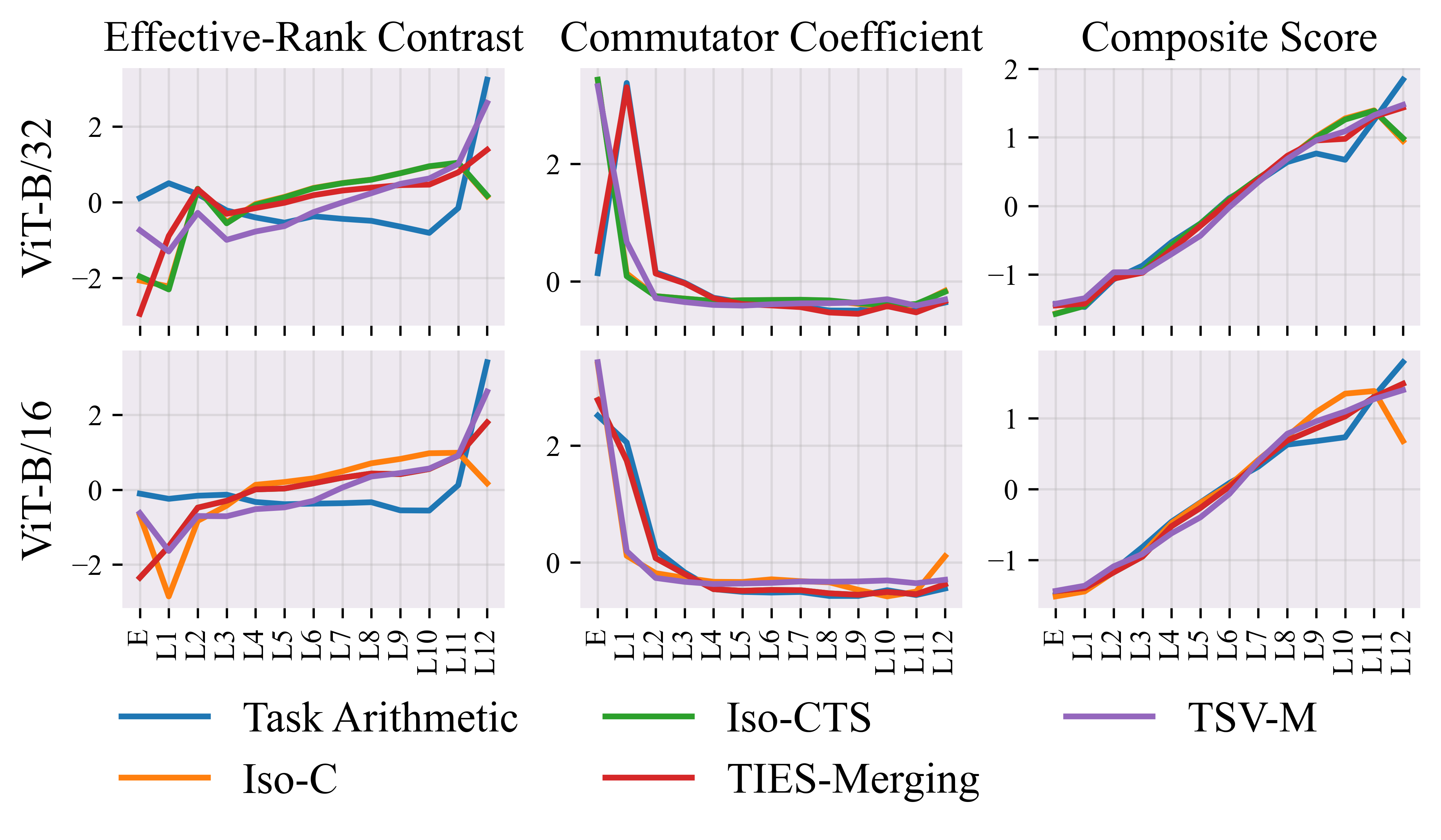}
    \caption{Layer-wise behavior of weight-only metrics. We show effective-rank contrast ($e_\ell$), commutator coefficient ($c_\ell$), and composite score ($s_\ell$) for five data-free merging methods on ViT-B/32 (top) and ViT-B/16 (bottom).
Deeper layers exhibit higher $e_\ell$ and lower $c_\ell$, producing 
monotonically increasing composite scores and motivating our depth-aware scaling rule.}
\vspace{3mm}
    \label{fig:weight-metrics}
\end{figure}

\paragraph{Continuous gate.}
We convert $w_\ell$ using a bounded gate function $r$:
\begin{equation*}
  s_\ell = r(w_\ell) = 1 + 0.5\,\tanh\!\big(\gamma\,(w_\ell-1)\big).
\end{equation*}
The $\tanh$ nonlinearity yields a smooth, saturating mapping centered at $1$ so small $w_\ell$ perturbations do not cause large changes, while bounding $s_\ell$ in $(0.5,1.5)$ prevents pathological over/under-scaling. Across all experiment we set $\gamma{=}3$. See the third column in Fig. \ref{fig:weight-metrics} where $s_\ell$ gradually increases with depth on ViT backbones that account for both information richness and conflict level. The additive constant in the denominator ensures identity scaling for a neutral layer, improves numerical stability, and performance is empirically insensitive to small variations of this offset.

\paragraph{Why bounded scaling and metric design.}
We constrain $s_\ell \in (0.5, 1.5)$ to avoid pathological scaling in the data-free regime. Under local smoothness along the merge direction, excess loss grows approximately quadratically with scaling deviation, making coarse bounded scaling robust to estimation noise. 
Effective-rank contrast $e_\ell$ favors concentrated, low-dimensional updates often associated with informative adaptation, while diffuse high-rank deltas resemble noise and are thus down-weighted. 
The commutator term $c_\ell$ measures operator non-commutativity; large values indicate feature-rotating updates that are more prone to interference and therefore benefit from shrinkage.

\paragraph{Tiered gate.}

Alternatively, we can map these raw weights to discrete layer scaling coefficients $s_\ell$ via a simple three-level schema. Rather than directly using $r_\ell$, which might vary continuously, we assign each layer into one of three categories: shallow, middle, or deep (top) layers, and we use a fixed scale value for each category. In our design, shallow layers receive a down-scaling (e.g. $s_\ell = 0.5$), middle layers use a neutral scaling ($s_\ell = 1.0$), and top layers receive an amplified scaling ($s_\ell = 1.5$).  Formally the schema can be defined as follows:
\begin{equation}
\label{eq:stier}
s^{\text{tier}}_\ell \in \{0.5,\,1.0,\,1.5\},
~
s^{\text{tier}}_\ell
=
\begin{cases}
0.5, & w_\ell \le t_1,\\[2pt]
1.0, & t_1 < w_\ell \le t_2,\\[2pt]
1.5, & w_\ell > t_2,
\end{cases}
\end{equation}
with fixed thresholds $(t_1,t_2)$ reused across models and mergers. This three-bucket assignment is determined by a non-linear mapping of the raw score and the layer index: for instance, lower-indexed layers tend to fall into the “shallow” bucket unless their $r_\ell$ is exceptionally high, whereas higher-indexed layers are biased toward the “deep” bucket. The exact boundaries for these buckets are fixed in advance as a part of our recipe (and kept constant for all models and tasks). 

To determine $(t_1,t_2)$, we define the empirical CDF $\widehat{F}_L(t)=\frac{1}{|\mathcal{A}|}\sum_{\ell\in\mathcal{A}}\mathbf{1}\{w_\ell\le t\}$ and set the tier thresholds as the $(\alpha,\beta)$ \emph{quantiles}:
\begin{equation*}
(t_1,t_2)\;=\;\Big(\ \inf\{t:\widehat{F}_L(t)\ge\alpha\},\ \inf\{t:\widehat{F}_L(t)\ge\beta\}\ \Big).
\label{eq:tier-quantiles}
\end{equation*}
Quantiles are scale-free and yield a stable fraction of layers in each bucket irrespective of the absolute range of $w_\ell$. 
Across tasks we choose to use $(\alpha,\beta)=(\tfrac{1}{3},\tfrac{2}{3})$ that gives roughly balanced terciles (shrink / neutral / amplify) and works well across backbones and mergers.

In essence, the combination of the metrics and $\tanh$ shaping yields an initial continuous estimate for how aggressively to apply the update at each layer, and then this step collapses that estimate to one of three discrete scaling levels (0.5, 1.0, or 1.5). This discretization adds a smoothing effect that avoids overfitting to idiosyncrasies of a particular model: it forces similar treatment for broad groups of layers (early vs. mid vs. late), which we found generalizes well across architectures. 

\paragraph{Depth prior versus weight-only diagnostics.}
Depth provides a useful coarse prior, but it cannot capture layer-specific structure and conflict. Linear or tiered depth schedules already improve over uniform scaling, yet they apply the same rule to all layers at a given depth. In contrast, the weight-only signals $e_\ell$ and $c_\ell$ introduce layer-adaptive modulation based on update structure and operator conflict, respectively. As shown in the ablation study, incorporating these diagnostics consistently outperforms depth-only scaling.

For completeness, we evaluate both the continuous composite-score scaling and the tiered scheme for every merge rule and backbone, and report the stronger of the two variants; a detailed discussion of the tiered search is provided in
the appendix. We emphasize that $r_\ell$ serves only as a simple depth prior with a modest
exponent, and is not intended to dominate the composite score; ablations in
Sec.~\ref{sec:ablation} show that the weight-only diagnostics $e_\ell$
and $c_\ell$ each provide additional gains beyond using $r_\ell$ alone.

\begin{algorithm}[t]
\caption{LARV: data-free layer-wise adaptive rescaling veneer over a base merger}
\label{alg:lavm}
\DontPrintSemicolon
\BlankLine
\hrule
\hrule
\BlankLine
\KwIn{Pretrained $\theta_0$; task deltas $\{\Delta\theta_{i,\ell}\}$; base merger $\mathcal{M}$; exponents $(\eta,\rho,\zeta)$; depth prior $p$; gate sharpness $\gamma$; \emph{mode} $\in\{\text{tiered},\text{continuous}\}$; tiered quantiles $(\alpha,\beta)$; small $\varepsilon>0$.}
\KwOut{Merged model $\theta_{\text{LARV}}$.}
\BlankLine
\hrule
\BlankLine

\For{$\ell=1$ \KwTo $L$}{
1.~Merge per-layer deltas.\;
  $\Delta\theta_{\ell}\leftarrow\mathcal{M}\big(\{\Delta\theta_{i,\ell}\}_{i=1}^{K}\big)$ \tcp*{Eq.~\eqref{eq:base-merge}}
  2.~Compute weight-only diagnostics.\;
  Calculate $e_\ell$ and  $c_\ell$ via  Eqs.~(\ref{eq:eer}-\ref{eq:ccc-rect}) \;
  3.~Composite weight-only scores.\;
Standardize to obtain $z(c_\ell)$ then \;
 $w_\ell \leftarrow \dfrac{e_\ell^{\,\eta}\; r_\ell^{\,\rho}}{\big(\mathrm{sp}(z(c_\ell))+1\big)^{\zeta}}$ \tcp*{Eq.~\eqref{eq:composite}}
 4A.~Tiered gate.\;
 
\If{\emph{mode} is \texttt{tiered}}{
    assign $s_\ell\in\{0.5,1.0,1.5\}$ by Eq.~\eqref{eq:stier} \;
}
\BlankLine
4B.~Continuous gate.\;
\If{\emph{mode} is \texttt{continuous}}{
    $\hat{s}_\ell \leftarrow 1 + 0.5\,\tanh\!\big(\gamma\,(w_\ell-1)\big)$ 
}
5.~Compose the merged model.\;
  $\theta_{\text{LARV},\,\ell} \leftarrow \theta_{0,\ell} + s_\ell\,\Delta\theta_\ell$ 
}
\Return $\theta_{\text{LARV}}$.
\BlankLine
\hrule
\hrule
\BlankLine
\end{algorithm}

\begin{table*}[t]
\centering
\small
\setlength{\tabcolsep}{6pt}
\renewcommand{\arraystretch}{0.7}

\begin{tabular}{@{}llllllllll@{}}
\toprule
\textbf{Method} & \textbf{SUN397} & \textbf{Cars} & \textbf{RESISC45} &
\textbf{EuroSAT} & \textbf{SVHN} & \textbf{GTSRB} & \textbf{MNIST} &
\textbf{DTD} & \textbf{Avg.} \\
\midrule


\multicolumn{1}{c}{} & \multicolumn{9}{c}{\textbf{ViT-B/32}} \\

\textit{fine-tuned} & 74.9 & 78.5 & 95.1 & 99.1 & 97.3 & 98.9 & 99.6 & 79.7 & 90.4 \\

\cmidrule(lr){2-10}

AdaMerging & 59.2 & 57.9 & 70.6 & 79.3 & 70.5 & 54.3 & --- & --- & 68.1 \\

Simple Averaging & 65.4 & 62.4 & 70.6 & 75.7 & 64.5 & 55.0 & 86.3 & 50.6 & 66.3 \\


\rowcolor{black!5}
\textbf{TA/LARV} &
60.1 \gainp{+3.1} &
64.9 \gainp{+9.2} &
73.3 \gainp{+8.5} &
87.9 \gainp{+14.6} &
81.1 \gainp{+3.2} &
76.9 \gainp{+8.4} &
97.3 \gainp{+1.2} &
55.9 \gainp{+8.7} &
74.7 \gainp{+7.1} \\


\rowcolor{black!5}
\textbf{ISO-C/LARV} &
70.9 \gainp{+1.2} &
75.6 \gainp{+2.7} &
90.0 \gainp{+4.0} &
\textbf{96.1} \gainp{+6.4} &
86.5 \gainm{-1.1} &
\textbf{94.6} \gainp{+2.2} &
98.8 \gainp{+0.3} &
74.4 \gainp{+9.0} &
\textbf{85.9} \gainp{+3.1} \\


\rowcolor{black!5}
\textbf{ISO-CTS/LARV} &
\textbf{72.9} \gainp{+1.8} &
\textbf{76.4} \gainp{+1.8} &
\textbf{90.3} \gainp{+3.8} &
93.4 \gainp{+4.3} &
81.6 \gainm{-1.8} &
91.4 \gainp{+1.0} &
98.3 \gainp{+0.2} &
\textbf{74.5} \gainp{+0.6} &
84.9 \gainp{+2.1} \\


\rowcolor{black!5}
\textbf{TIES/LARV} &
68.7 \gainp{+1.7} &
67.9 \gainp{+3.7} &
79.6 \gainp{+5.3} &
84.9 \gainp{+10.4} &
77.8 \gainp{+0.0} &
76.3 \gainp{+6.9} &
95.4 \gainp{+1.3} &
58.9 \gainp{+4.9} &
76.2 \gainp{+4.3} \\


\rowcolor{black!5}
\textbf{TSV-M/LARV} &
69.9 \gainp{+2.3} &
74.9 \gainp{+3.2} &
88.5 \gainp{+3.8} &
95.6 \gainp{+2.2} &
\textbf{90.1} \gainm{-1.8} &
92.7 \gainp{+0.2} &
\textbf{98.9} \gainp{+0.0} &
70.9 \gainp{+7.0} &
85.2 \gainp{+2.1} \\

\midrule


\multicolumn{1}{c}{} & \multicolumn{9}{c}{\textbf{ViT-B/16}} \\

\textit{fine-tuned} & 78.9 & 85.9 & 96.6 & 99.1 & 97.6 & 99.0 & 99.7 & 82.3 & 92.3 \\

\cmidrule(lr){2-10}

AdaMerging & 67.1 & 65.7 & 78.7 & 86.0 & 86.8 & 91.6 & 96.4 & 45.2 & 45.2 \\

Simple Averaging & 68.7 & 69.0 & 75.1 & 83.3 & 75.0 & 62.6 & 93.8 & 51.2 & 72.3 \\


\rowcolor{black!5}
\textbf{TA/LARV} &
66.1 \gainp{+0.2} &
71.6 \gainp{+3.3} &
78.1 \gainp{+2.6} &
86.7 \gainp{+2.2} &
89.4 \gainp{+0.6} &
84.3 \gainp{+2.3} &
98.6 \gainp{+0.5} &
58.1 \gainp{+4.1} &
79.1 \gainp{+2.0} \\


\rowcolor{black!5}
\textbf{ISO-C/LARV} &
\textbf{76.7} \gainp{+0.9} &
\textbf{85.3} \gainp{+2.5} &
93.2 \gainp{+0.9} &
96.9 \gainp{+0.6} &
91.0 \gainm{-0.8} &
94.9 \gainm{-0.1} &
98.8 \gainm{-0.1} &
\textbf{76.8} \gainp{+4.8} &
\textbf{89.2} \gainp{+1.1} \\


\rowcolor{black!5}
\textbf{ISO-CTS/LARV} &
75.5 \gainm{-0.6} &
83.2 \gainm{-0.6} &
\textbf{93.4} \gainp{+0.8} &
\textbf{97.2} \gainp{+1.2} &
\textbf{93.4} \gainp{+2.5} &
\textbf{96.6} \gainp{+1.9} &
98.9 \gainp{+0.3} &
74.7 \gainp{+1.0} &
89.1 \gainp{+0.8} \\


\rowcolor{black!5}
\textbf{TIES/LARV} &
72.2 \gainp{+1.5} &
73.2 \gainp{+2.0} &
84.8 \gainp{+4.9} &
90.6 \gainp{+3.1} &
90.9 \gainp{+7.6} &
87.6 \gainp{+11.3} &
98.3 \gainp{+1.9} &
61.1 \gainp{+5.6} &
82.3 \gainp{+4.7} \\


\rowcolor{black!5}
\textbf{TSV-M/LARV} &
74.0 \gainp{+0.9} &
82.6 \gainp{+1.9} &
91.3 \gainp{+1.6} &
96.6 \gainp{+0.4} &
93.1 \gainm{-1.0} &
93.7 \gainm{-0.4} &
\textbf{99.1} \gainp{+0.0} &
75.4 \gainp{+5.7} &
88.2 \gainp{+1.1} \\

\midrule


\multicolumn{1}{c}{} & \multicolumn{9}{c}{\textbf{ViT-L/14}} \\

\textit{fine-tuned} & 82.8 & 92.8 & 97.4 & 99.1 & 97.9 & 99.2 & 99.8 & 85.5 & 94.3 \\

\cmidrule(lr){2-10}

AdaMerging & 75.9 & 80.2 & 78.1 & 82.0 & 68.3 & 93.2 & 93.1 & 68.6 & 79.9 \\

Simple Averaging & 72.5 & 81.5 & 82.3 & 88.5 & 81.6 & 74.0 & 96.6 & 61.8 & 79.9 \\


\rowcolor{black!5}
\textbf{TA/LARV} &
74.5 \gainp{+2.5} &
83.3 \gainp{+4.3} &
86.2 \gainp{+5.6} &
92.7 \gainp{+8.1} &
88.9 \gainp{+1.4} &
88.8 \gainp{+5.3} &
98.6 \gainp{+0.6} &
63.8 \gainp{+5.3} &
84.6 \gainp{+4.1} \\


\rowcolor{black!5}
\textbf{ISO-C/LARV} &
80.8 \gainp{+0.1} &
92.4 \gainp{+0.9} &
95.9 \gainp{+0.5} &
97.5 \gainp{+0.3} &
94.6 \gainm{-0.6} &
\textbf{97.8} \gainp{+0.0} &
\textbf{99.2} \gainp{+0.1} &
82.9 \gainp{+2.6} &
92.6 \gainp{+0.5} \\


\rowcolor{black!5}
\textbf{ISO-CTS/LARV} &
\textbf{81.4} \gainp{+0.3} &
\textbf{93.0} \gainp{+0.8} &
\textbf{96.4} \gainp{+0.8} &
97.6 \gainp{+0.3} &
94.1 \gainm{-0.5} &
97.3 \gainm{-0.2} &
\textbf{99.2} \gainp{+0.1} &
\textbf{83.6} \gainp{+2.5} &
\textbf{92.8} \gainp{+0.5} \\


\rowcolor{black!5}
\textbf{TIES/LARV} &
77.3 \gainp{+2.6} &
86.7 \gainp{+3.5} &
91.1 \gainp{+4.6} &
94.4 \gainp{+4.7} &
88.9 \gainm{-0.8} &
88.8 \gainp{+3.6} &
98.2 \gainp{+0.4} &
68.7 \gainp{+4.8} &
86.7 \gainp{+2.9} \\


\rowcolor{black!5}
\textbf{TSV-M/LARV} &
79.8 \gainp{+1.6} &
91.0 \gainp{+1.2} &
94.4 \gainp{+0.9} &
\textbf{97.8} \gainp{+1.1} &
\textbf{95.2} \gainm{-0.4} &
96.9 \gainp{+0.4} &
\textbf{99.2} \gainp{+0.1} &
79.5 \gainp{+4.2} &
91.7 \gainp{+1.1} \\

\bottomrule
\end{tabular}

\caption{ViT results on 8 classification tasks for ViT-B/32, ViT-B/16, and ViT-L/14. 
All numbers are percentages. For each LARV variant, the gray parenthetical value 
shows its performance gain relative to the corresponding baseline, allowing the 
baseline score to be implicitly recovered. Blue indicates a positive gain, while 
red denotes a decrease in accuracy.}
\label{tab:b32_8tasks}
\end{table*}

\subsubsection{Discussion}
Putting these components together, Algorithm~\ref{alg:lavm} summarizes our proposed veneer, which can be attached as a training-free add-on to any base merger $\mathcal{M}$. The layer-wise gating is merger-agnostic and data-free, so the same fixed layer-scaling policy is used across all experiments. Empirically, LARV’s layer-wise adaptation consistently preserves accuracy better than uniform merging, indicating robustness and broad applicability.

Conceptually, our method offers a unified view that recovers uniform merging as a special case ($s_1{=}\cdots{=}s_L$) while also admitting optional unlabeled probes without any additional learning. It further refines the understanding of model merging by probing the layer-wise structure through two complementary lenses: $e_\ell$ rewards concentrated updates that stay aligned with base semantics (akin to successful low-rank adaptation~\cite{hu2021lora}), and $c_\ell$ penalizes order-sensitive interference, a notion complementary to permutation-based alignment~\cite{ainsworth2023gitrebasin,entezari2021lmc} and orthogonal to Fisher reweighting and sign-trimming rules~\cite{matena2022merging,yadav2023ties}.

\subsection{Complexity and Implementation} 
\label{sec:method:complexity}
LARV introduces computational overhead mainly affected by lightweight randomized SVD. $O(m_\ell n_\ell k L)$ operations where $m_\ell$ and $n_\ell$ and the dimensions of each layer's matrices and $k$ is the rank of the matrices, for computing $(e_\ell, c_\ell)$ and applying per-layer scaling.
No additional gradients or memory copies are required.
The implementation can be completed in under 20 lines of Python and integrates seamlessly into the FusionBench~\cite{tang2024fusionbench} framework.

\paragraph{No merge-time tuning.}
Although LARV introduces coefficients $(\eta, \rho, \zeta, \gamma)$, these are not merge-time tuning knobs. All experiments use one fixed configuration across tasks, backbones, and base mergers, with no gradients, validation sets, or task-specific search. Sensitivity analysis (Appendix) shows performance remains stable across wide parameter ranges, indicating that LARV operates in a robust, no-tuning regime.

\section{Results and Analysis}
\label{sec:results}

\subsection{Experimental Setup}
We evaluate \textbf{LARV merging} on CLIP ViT~\cite{radford2021learning} backbones \textbf{B/32}, \textbf{B/16}, and \textbf{L/14} under three merge settings~\cite{huang2024emr,lu2024twin} with \textbf{8 Tasks}, \textbf{14 Tasks}, and \textbf{20 Tasks} classification tasks using FusionBench~\cite{tang2024fusionbench}.
For the 8-task suite we report per-dataset scores on
\textbf{SUN397}~\cite{xiao2010sun}, \textbf{Stanford Cars}~\cite{krause20133d}, \textbf{RESISC45}~\cite{cheng2017remote}, \textbf{EuroSAT}~\cite{helber2019eurosat}, \textbf{SVHN}~\cite{netzer2011reading}, \textbf{GTSRB}~\cite{stallkamp2011german}, \textbf{MNIST}~\cite{lecun1998mnist}, and \textbf{DTD}~\cite{cimpoi2014describing}.
We compare to representative training-free baselines: \emph{Simple Averaging}, \emph{Task Arithmetic (TA)}, \emph{ISO-C} and \emph{ISO-CTS}~\cite{marczak2025no} , \emph{TIES-Merging}~\cite{yadav2023ties}, and \emph{TSV-M}~\cite{gargiulo2025task}.
LARV is used as a \emph{veneer} on each base rule (e.g., ``TIES/LARV''): the only change is replacing the uniform scale in Eq.~(2) with the layer-wise scales in Eq.~(3).
No data and no retraining are used anywhere.
All numbers are top-1 accuracy (\%) with the average taken over tasks in each block.
Parentheses in LARV rows denote the absolute gain over the corresponding base rule.

\begin{table*}[t]
\centering
\small
\setlength{\tabcolsep}{6pt}
\renewcommand{\arraystretch}{0.8}
\begin{tabular}{@{}lccc|ccc|ccc@{}}
\toprule
& \multicolumn{3}{c|}{\textbf{ViT-B/32}} 
& \multicolumn{3}{c|}{\textbf{ViT-B/16}} 
& \multicolumn{3}{c}{\textbf{ViT-L/14}} \\
\textbf{Method} 
& \textbf{8 Tasks} & \textbf{14 Tasks} & \textbf{20 Tasks}
& \textbf{8 Tasks} & \textbf{14 Tasks} & \textbf{20 Tasks}
& \textbf{8 Tasks} & \textbf{14 Tasks} & \textbf{20 Tasks} \\
\midrule
Pretrained 
& 48.3 & 57.2 & 56.1
& 55.3 & 61.3 & 59.7
& 64.7 & 68.2 & 65.2 \\
Fine-tuned 
& 92.8 & 90.9 & 91.3
& 94.6 & 92.8 & 93.2
& 95.8 & 94.3 & 94.7 \\
\midrule
\rowcolor{black!5}\textbf{TA/LARV}
& 73.8 \gainp{+6.3} & 65.8 \gainp{+13.0} & 83.3 \gainp{+19.2}
& 79.1 \gainp{+2.0} & 73.3 \gainp{+12.5} & 88.6 \gainp{+23.7}
& 84.6 \gainp{+4.1} & 76.3 \gainp{+13.2} & 55.0 \gainp{+19.0} \\

\rowcolor{black!5}\textbf{ISO-C/LARV}
& \textbf{85.9} \gainp{+3.1} & 81.2 \gainp{+3.1} & 93.9 \gainp{+2.6}
& \textbf{89.2} \gainp{+1.1} & 84.3 \gainp{+1.9} & 96.1 \gainp{+1.2}
& 92.6 \gainp{+0.5} & 90.5 \gainp{+1.0} & 85.4 \gainp{+2.0} \\

\rowcolor{black!5}\textbf{ISO-CTS/LARV}
& 84.9 \gainp{+2.2} & \textbf{82.4} \gainp{+3.0} & \textbf{94.6} \gainp{+2.2}
& 89.1 \gainp{+0.8} & \textbf{86.1} \gainp{+1.6} & 96.1 \gainp{+0.9}
& \textbf{92.8} \gainp{+0.5} & \textbf{91.1} \gainp{+0.7} & \textbf{87.3}  \gainp{+1.5} \\

\rowcolor{black!5}\textbf{TIES/LARV}
& 76.3 \gainp{+4.4} & 73.6 \gainp{+6.0} & 91.5 \gainp{+5.2}
& 82.3 \gainp{+4.7} & 77.3 \gainp{+5.8} & 94.7 \gainp{+4.7}
& 86.7 \gainp{+2.9} & 83.7 \gainp{+5.9} & 70.4 \gainp{+7.4} \\

\rowcolor{black!5}\textbf{TSV-M/LARV}
& 83.1 \gainp{+6.6} & 81.2 \gainp{+2.4} & \textbf{94.6} \gainp{+1.3}
& 88.2 \gainp{+1.1} & 84.2 \gainp{+2.1} & \textbf{96.2} \gainp{+1.1}
& 91.7 \gainp{+1.1} & 89.8 \gainp{+1.6} & 86.4 \gainp{+3.2} \\
\bottomrule
\end{tabular}
\caption{\textbf{Performance summary on 8/14/20 tasks.} 
Average accuracy (\%) for each ViT backbone across 8/14/20-task merges. 
Blue numbers denote the absolute improvement of each LARV variant over the corresponding baseline without LARV. 
The results show that LARV consistently boosts all merging rules, with larger gains appearing on smaller backbones (ViT-B/32, ViT-B/16) and more challenging merge settings (20 tasks).}

\label{tab:9cells}
\end{table*}
\begin{table}[t]
\centering
\small
\setlength{\tabcolsep}{6pt}
\renewcommand{\arraystretch}{0.8}
\begin{tabular}{ccc|cccc}
\toprule
$\mathbf{r_\ell}$ & $\mathbf{e_\ell}$ & $\mathbf{c_\ell}$ & \textbf{TIES} & \textbf{Iso-C} & \textbf{Iso-CTS} & \textbf{TSV-M} \\
\midrule
– & – & –  & 0.719 & 0.828 & 0.827 & 0.831 \\
– & – & \checkmark  & 0.734 & 0.833 & 0.833 & 0.830 \\
\checkmark & – & –  & 0.751 & 0.847 & 0.851 & 0.846 \\
\checkmark & – & \checkmark  & 0.756 & 0.848 & 0.851 & 0.846 \\
\checkmark & \checkmark & \checkmark  & 0.759 & 0.850 & 0.854 & 0.850 \\
\bottomrule
\end{tabular}
a\caption{Ablation on layer metrics $r_\ell, e_\ell, c_\ell$ and their effect on several merging baselines.}
\label{tab:metric-ablation}
\end{table}
\subsection{Main Results Across Backbones (8 tasks)}
Table~\ref{tab:b32_8tasks} presents the full 8-task results for ViT-B/32,
ViT-B/16, and ViT-L/14.  
Across all three backbones and all merge rules, applying the LARV veneer leads
to a consistent improvement over the underlying baseline.  
The magnitude of the gain varies with the strength of the base rule, but the
trend is uniform: weaker methods such as TA and TIES benefit the most, while
stronger baselines such as Iso-C, Iso-CTS, and TSV-M obtain smaller yet
reliable boosts.

A clear pattern appears across backbones.  
LARV produces its largest improvements on datasets where layer-wise
heterogeneity is more pronounced, such as \textit{RESISC45}, \textit{EuroSAT},
and \textit{GTSRB}.  
These tasks rely more heavily on deeper visual representations, and the
depth-dependent scaling provided by LARV helps stabilize the merge and reduce
cross-layer conflict.  
Meanwhile, high-accuracy datasets such as \textit{MNIST} and \textit{SVHN}
remain near ceiling, indicating that the veneer does not disturb layers that
already align well.

\paragraph{Key Observations.}
(1) LARV improves every backbone and every merge rule, with gains ranging from
small but consistent (Iso-CTS, TSV-M) to substantial (TA, TIES).  
(2) Improvements concentrate on tasks that depend more strongly on mid-to-deep
layers, consistent with the intended effect of LARV’s depth-aware scaling.  
(3) Performance is stable across model scales (B/32, B/16, L/14), suggesting that
the three-signal layer scaling generalizes reliably as backbone capacity grows.

\subsection{Summary Across Backbones and Task Scales}

Table~\ref{tab:9cells} summarizes average accuracies for the three ViT backbones under 8-, 14-, and 20-task merges. 
Each cell reports the merged accuracy and the absolute gain achieved by attaching
the LARV veneer to the corresponding baseline. 
Across all settings, LARV consistently improves every base rule, including 
Task Arithmetic, ISO-C/CTS, TIES-Merging, and TSV-M.

\textbf{Scale across model size.}
Gains are largest on smaller backbones (e.g., ViT-B/32), which are more sensitive 
to cross-task interference and therefore benefit more from suppressing 
shallow-layer noise and amplifying deeper-layer structure.

\textbf{Scale with merge difficulty.}
LARV brings noticeably larger gains in more challenging scenarios, especially in the 20-task experiments where interference between models is strongest and uniform scaling often falls short. This pattern indicates that layer-wise adaptation becomes increasingly important as the number of merged tasks grows.

Taken together, the results show that LARV scales well with both model capacity (from B/32 to L/14) and merge difficulty (from 8 to 20 tasks), delivering consistent improvements across architectures without requiring data or extra tuning.

\begin{table*}[t]
\centering
\small
\setlength{\tabcolsep}{3pt}
\renewcommand{\arraystretch}{0.8}
\begin{tabular}{@{}lllllllllll@{}}
\toprule
\textbf{Method} &
\multicolumn{7}{c}{\textbf{Seen Tasks}} &
\multicolumn{3}{c}{\textbf{Unseen Tasks}} \\
\cmidrule(lr){2-8}
\cmidrule(lr){9-11}
& SUN397 & Cars & RESISC45 & DTD & SVHN & GTSRB & Avg. &
MNIST & EuroSAT & Avg. \\
\midrule


\multicolumn{1}{c}{} &
\multicolumn{10}{c}{ViT-B/32 Generalization} \\

\textit{Pre-trained} &
63.2 & 59.9 & 60.6 & 43.9 & 23.5 & 30.4 & 46.9 &
47.6 & 45.6 & 46.6 \\

\cmidrule(lr){2-11}

Fisher Merging &
65.5 & 67.2 & 78.2 & 57.6 & 84.2 & 75.9 & 71.4 &
71.8 & 49.4 & 60.6 \\

RegMean &
69.5 & 70.8 & 88.7 & 67.2 & 95.2 & 89.4 & 80.1 &
82.9 & 44.6 & 63.8 \\

RegMean++ &
69.8 & 70.8 & 90.2 & 70.3 & \textbf{95.5} & \textbf{93.2} & 81.6 &
81.3 & 44.1 & 62.7 \\



Layer-wise AdaMerging &
68.4 & 71.9 & 87.9 & 69.1 & 92.2 & 93.8 & 80.5 &
77.7 & 47.3 & 62.5 \\



\rowcolor{black!5}
\textbf{TSV-M/LARV} &
70.1 \gainp{+1.0} &
75.5 \gainp{+2.7} &
91.6 \gainp{+3.3} &
71.2 \gainp{+4.6} &
94.1 \gainm{-0.4} &
95.1 \gainm{-0.2} &
\textbf{82.9} \gainp{+1.8} &
\textbf{86.2} \gainp{+1.8} &
49.6 \gainp{+3.0} &
67.9 \gainp{+2.2} \\

\rowcolor{black!5}
\textbf{TA/LARV} &
65.9 \gainp{+1.7} &
67.2 \gainp{+4.2} &
79.7 \gainp{+6.5} &
59.6 \gainp{+4.9} &
86.2 \gainp{+1.5} &
84.0 \gainp{+4.5} &
73.8 \gainp{+3.9} &
78.4 \gainp{+3.0} &
43.2 \gainp{+2.5} &
60.8 \gainp{+2.7} \\

\rowcolor{black!5}
\textbf{ISO-C/LARV} &
\textbf{74.4} \gainp{+2.3} &
78.3 \gainp{+3.9} &
92.2 \gainp{+4.2} &
73.5 \gainp{+5.4} &
85.2 \gainm{-0.5} &
93.2 \gainp{+1.4} &
82.8 \gainp{+2.8} &
80.9 \gainp{+2.3} &
56.4 \gainp{+3.3} &
\textbf{68.6} \gainp{+2.7} \\

\rowcolor{black!5}
\textbf{ISO-CTS/LARV} &
74.2 \gainp{+2.5} &
\textbf{78.4} \gainp{+4.1} &
\textbf{92.5} \gainp{+4.7} &
\textbf{75.0} \gainp{+5.1} &
83.8 \gainp{+0.3} &
93.5 \gainp{+1.7} &
82.9 \gainp{+3.1} &
79.0 \gainp{+1.9} &
\textbf{57.4} \gainp{+5.0} &
68.2 \gainp{+3.4} \\

\rowcolor{black!5}
\textbf{TIES/LARV} &
69.9 \gainp{+1.8} &
67.8 \gainp{+2.1} &
81.6 \gainp{+5.1} &
59.1 \gainp{+4.2} &
76.3 \gainp{+0.9} &
77.8 \gainp{+5.7} &
72.1 \gainp{+3.3} &
74.1 \gainp{+1.1} &
49.5 \gainp{+2.6} &
61.8 \gainp{+1.8} \\

\bottomrule
\end{tabular}
\caption{Generalization results for ViT-B/32 on both seen and unseen tasks.
All numbers are percentages. Parenthetical values indicate performance gain
relative to the corresponding baseline without LARV. LARV consistently improves
in-distribution accuracy while also enhancing cross-task generalization to
unseen datasets (MNIST, EuroSAT) without any retraining.}
\label{tab:id_ood}
\end{table*}


\subsection{Ablation on Weight-Only Metrics ($r_\ell$, $e_\ell$, $c_\ell$)}
\label{sec:ablation}

We ablate the three weight-only components used in LARV’s composite score:
the depth prior $r_\ell$, the information signal $e_\ell$, and the conflict
signal $c_\ell$. Table~\ref{tab:metric-ablation} evaluates all valid
combinations of these terms across four representative baselines:
TIES-Merging, ISO-C, ISO-CTS, and TSV-M.

Across all methods, the trend is consistent.  
Using none of the signals yields the weakest performance, confirming that
layer-wise variability is essential for effective merging.  
Adding the depth prior $r_\ell$ alone already recovers a substantial portion
of the overall gain, reflecting the broad depth-related structure in ViT
updates.  
Introducing $e_\ell$ further improves accuracy by elevating layers whose
updates contain more structured, informative variation.  
The conflict term $c_\ell$ provides an additional refinement by suppressing
non-commutative, interference-prone updates.

When all three terms $(r_\ell, e_\ell, c_\ell)$ are used together, each
baseline achieves its strongest or near-strongest result.  
This demonstrates that LARV is not simply a hand-crafted depth schedule:
$r_\ell$ explains part of the improvement, but both $e_\ell$ and $c_\ell$
contribute complementary information that cannot be recovered from depth
alone.  
Overall, the full combination yields the most stable and effective
layer-wise scaling.

\begin{figure}[t]
    \centering
    \includegraphics[width=\linewidth]{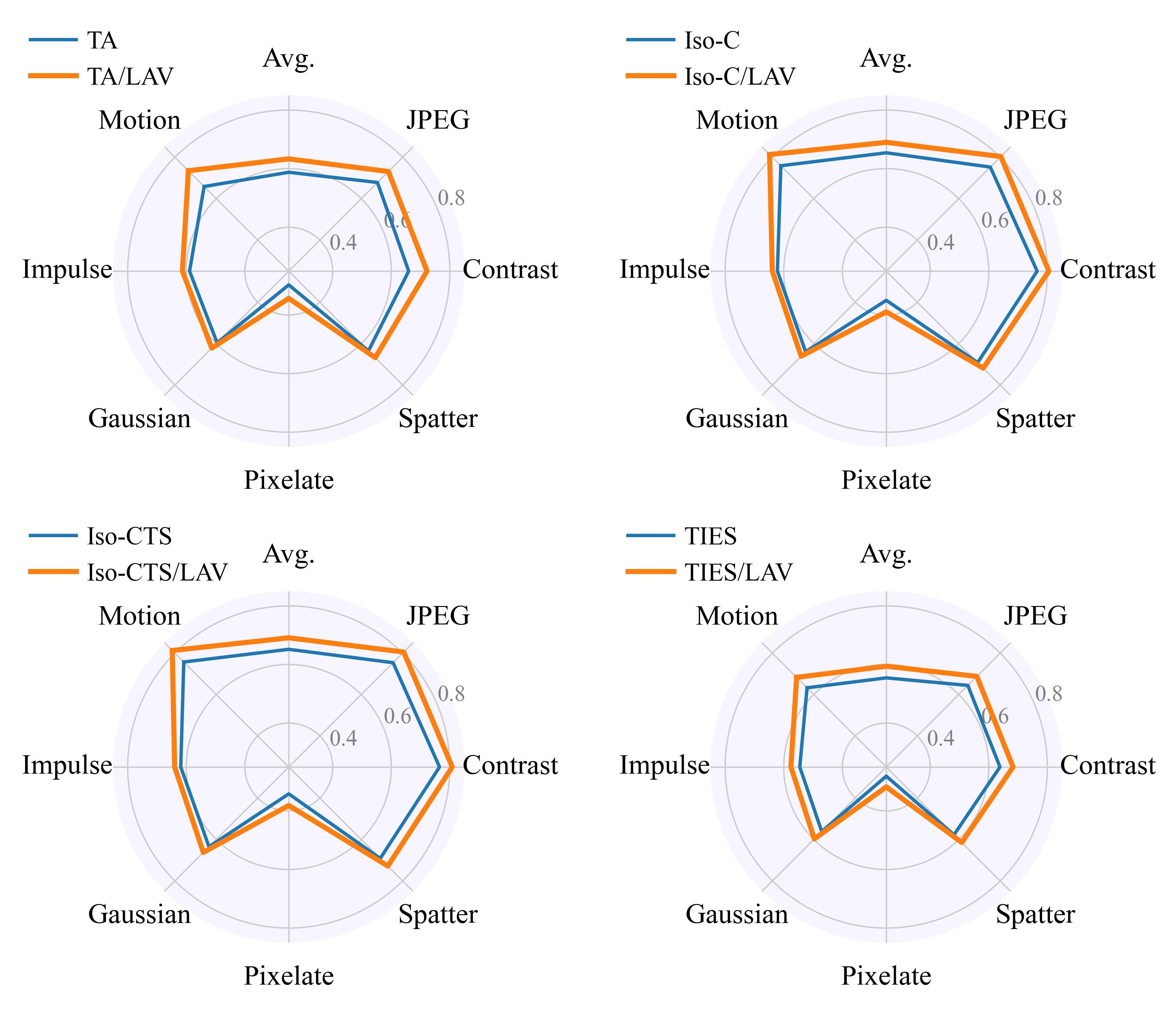}
    \caption{
        Performance on 7 corruption types and overall average.
        LARV consistently improves robustness across all corruption methods.
    }
    \vspace{-3mm}
    \label{fig:corruption}
\end{figure}

\subsection{Robustness under Input Corruptions}
\label{sec:corruption}
We further evaluate robustness using a set of common input corruptions
(blurring, noise, and digital artifacts), following the ImageNet-C convention\cite{hendrycks2019benchmarking}.
For each corruption type we measure accuracy across severities and report the
mean in Fig.~\ref{fig:corruption}.  Across all four merge rules, applying LARV
yields a consistent improvement over the corresponding baseline.  The gains are
most visible on corruptions that strongly affect low-level structure, such as
Motion, Impulse, and Gaussian noise, where suppressing shallow-layer responses
reduces the propagation of spurious activations.  Improvements are also observed
on more feature-oriented corruptions such as Contrast and JPEG, indicating that
the depth-dependent scaling helps preserve mid- and high-level representations
even when the input distribution is perturbed.

The effect is uniform across rules: TA, Iso-C, Iso-CTS, and TIES each benefit
from LARV, and no degradation is observed on any corruption type.  This
stability is noteworthy because the veneer is applied without retraining or
corruption-aware tuning.  Combined with the clean-data results, these findings
demonstrate that the proposed layer-wise scaling not only mitigates cross-layer
interference but also enhances resilience under distribution shift.

\subsection{Generalization to Unseen Tasks}

We further assess whether the benefits of LARV extend beyond the task suite used
to construct the task-vectors.  
Table~\ref{tab:id_ood} reports results for ViT-B/32 on eight seen
tasks and two additional unseen evaluation datasets.  The comparison includes
Fisher Merging, RegMean, RegMean++, and layer-wise AdaMerging.

Across all baselines, adding the LARV veneer consistently improves performance
on the seen tasks, with notable gains on RESISC45, DTD, and GTSRB.  These
improvements also transfer to the unseen datasets: LARV-enhanced variants obtain
higher accuracy on MNIST and EuroSAT without any retraining or domain-specific
adjustments.  
The consistency of these gains indicates that LARV reduces cross-layer
interference in a manner that generalizes across datasets, rather than relying
on characteristics specific to the tasks used during merge construction.

\section{Conclusion}
\label{sec:conclusion}

We introduced LARV, a data-free, training-free layer-wise rescaling veneer that improves model merging by replacing a single global coefficient with depth-aware scales derived from two simple weight-only diagnostics, information richness $e_\ell$ and conflict level $c_\ell$. Across all ViT backbones and merge sizes, LARV consistently enhances Task Arithmetic, ISO-C/CTS, TIES-Merging, and TSV-M, with the largest gains appearing on smaller models and harder 20-task merges. It further improves robustness under input corruptions and generalizes to unseen tasks without any tuning. These results show that acknowledging depth heterogeneity is an effective inductive bias for training-free merging, turning task-vector composition into a more stable and structurally informed procedure.

{
    \small
    \bibliographystyle{ieeetr}
    \bibliography{references}
}
\clearpage

\appendix 
\begin{center}
{\Large \textbf{Supplementary Material}}
\end{center}

\section{Formal Definitions and Default Hyperparameters}
\subsection{Definitions of $r_\ell$, $r(\cdot)$, and $\zeta$}
\label{supp:definitions}

In section \ref{sec:composite-gate} We refer the depth prior $r_\ell$, the gate mapping $r(\cdot)$, and $\zeta$ in compositing weight-only score and gate.
For completeness and reproducibility, we provide the formal
definitions of $r_\ell$ and  $\zeta$ used in the continuous variant, and the exponent
$\zeta$ in the composite score of Eq.~(8).

\paragraph{Depth prior $r_\ell$.}
We define $r_\ell$ as the normalized layer index:
\[
r_\ell = \frac{\ell}{L},
\]
where $\ell \in \{1,\dots,L\}$ and $L$ is the total number of
parameterized layers. Thus $r_\ell$ increases monotonically
from shallow to deep layers and remains independent of any
task or model.

The quantity $r_\ell$ provides a mild monotonic depth prior
that reflects the empirical observation that deeper layers tend
to exhibit more stable and semantically meaningful updates.
It is not intended to dominate the composite score; instead,
the weight-only diagnostics $e_\ell$ and $c_\ell$ contribute the
majority of the layer-wise variation, as confirmed by the
ablations in Table~\ref{tab:metric-ablation}.

\paragraph{Gate function $r(\cdot)$ in the continuous variant.}
In the continuous version of LARV, the raw composite score
$w_\ell$ is mapped to a bounded scaling coefficient $s_\ell$
via the gate function
\[
s_\ell = r(w_\ell)
= 1 + 0.5 \tanh\!\big(\gamma ( w_\ell - 1 )\big),
\]
where $\gamma > 0$ controls sharpness. Throughout all experiments we set $\gamma = 3$ for the continuous
variant. Note that $\gamma$ is not used in the tiered version,
which directly discretizes $w_\ell$ using the $(\alpha,\beta)$
quantiles instead of applying the smooth gate.

\paragraph{Exponent $\zeta$.}
The exponent $\zeta$ balances the influence of the
conflict-related term $(sp(z(c_\ell)) + 1)$ within the
product-of-experts composite score:
\[
w_\ell = e_\ell^\eta \, r_\ell^\rho \, (sp(z(c_\ell)) + 1)^\zeta.
\]
We fix $\zeta = 1$ for all experiments. This keeps the
conflict penalty linear in the standardized commutator score
and avoids over-emphasizing noisier layers.

\paragraph{Default hyperparameters.}
Unless otherwise specified, we use the configuration
$(\eta,\rho,\zeta) = (1, 0.5, 1)$ across all backbones,
task counts, and merge rules. These values were chosen for
simplicity and robustness and were not tuned per model.


\section{Additional Details on Complexity}
\label{supp:complexity}
We elaborate the details of complexity in  section \ref{sec:method:complexity}.
\paragraph{Randomized SVD setup.}
For each layer $\ell$, LARV computes two quantities $(e_\ell, c_\ell)$ based on
a lightweight randomized SVD of the weight difference matrix
$W_\ell \in \mathbb{R}^{m_\ell \times n_\ell}$.
Let $k$ denote the target rank and let $\alpha$ denote the oversampling
parameter used in the randomized sketch (typically $\alpha \in [5, 20]$).
The sketches therefore operate on a subspace of dimension
$k' = k + \alpha$.

\paragraph{Computational complexity.}
The standard complexity of a single randomized SVD is
\[
O(m_\ell n_\ell k') + O(n_\ell {k'}^{2}),
\]
where the first term corresponds to the randomized projection
and the second to the QR/SVD of the sketched matrix.
Because $k' \ll \min(m_\ell, n_\ell)$ in all practical settings,
the first term dominates and yields
\[
O(m_\ell n_\ell (k + \alpha)).
\]

\paragraph{Simplified expression used in the main paper.}
Since $\alpha$ is a small constant independent of model size
and $k$ is fixed across layers, we simplify the expression in
the main paper as
\[
O(m_\ell n_\ell k L),
\]
where $L$ is the number of layers.
This notation captures the correct scaling with respect to
matrix size and number of layers while omitting lower-order
constants $(\alpha)$ for clarity.

\paragraph{Choice of $k$ and $\alpha$.}
In all experiments we set $k \in \{1, 2\}$ depending on the
matrix shape, and we use a constant oversampling
$\alpha = 10$ following standard recommendations in
randomized numerical linear algebra.
We do not perform power iterations, so the above complexity
is tight.
We emphasize that LARV does not require backpropagation,
gradient computation, or additional memory copies; the
randomized SVD is applied once per layer and the resulting
scaling factors are stored as scalars.

\section{Experimental Setup and Protocol}
\label{app:setup}



\begin{table*}[t]
\centering
\small
\setlength{\tabcolsep}{6pt}
\renewcommand{\arraystretch}{1.0}

\begin{tabular}{@{}l|ccc|ccc|ccc|ccc@{}}
\toprule
& \multicolumn{3}{c|}{\textbf{TIES}} 
& \multicolumn{3}{c|}{\textbf{ISO-C}} 
& \multicolumn{3}{c|}{\textbf{ISO-CTS}} 
& \multicolumn{3}{c}{\textbf{TSV-M}} \\
\midrule
\textbf{Strategy} 
& \textbf{B/32} & \textbf{B/16} & \textbf{L/14}
& \textbf{B/32} & \textbf{B/16} & \textbf{L/14}
& \textbf{B/32} & \textbf{B/16} & \textbf{L/14}
& \textbf{B/32} & \textbf{B/16} & \textbf{L/14} \\
\midrule

Uniform (basekube)  
& 71.9 & 77.6 & 83.8
& 82.8 & 88.1 & 92.1
& 82.7 & 88.3 & 92.3
& 83.1 & 87.1 & 90.6 \\

Linear
& 75.9 & 79.1 & 85.7
& 85.2 & 88.6 & 92.4
& 85.5 & 89.1 & 92.8
& 85.0 & 88.1 & 91.6 \\

Composite Score
& \textbf{77.4} & 79.2 & \textbf{86.6}
& \textbf{85.7} & 79.3 & 87.5
& \textbf{86.0} & 79.5 & 87.9
& 84.8 & 87.7 & 91.6 \\

Tier 12
& 75.8 & 79.1 & 85.7
& 85.1 & 88.6 & 92.4
& 85.5 & 89.1 & \textbf{92.8}
& 85.0 & 88.1 & 91.6 \\

Tier 6
& 76.0 & 79.2 & 85.9
& 85.3 & 88.7 & 92.4
& 85.7 & 89.1 & \textbf{92.8}
& 85.1 & 88.2 & 91.7 \\

Tier 3
& 76.3 & \textbf{79.3} & 86.1
& 85.4 & \textbf{88.8} & \textbf{92.5}
& 85.6 & \textbf{89.2} & \textbf{92.8}
& \textbf{85.1} & \textbf{88.2} & \textbf{91.7} \\

Tier 2
& 76.2 & 79.2 & 86.2
& 85.5 & 88.4 & 92.3
& 85.8 & 88.9 & 92.6
& 84.8 & 87.7 & 91.6 \\

\bottomrule
\end{tabular}

\caption{
Layer-wise scaling strategy comparison across ViT-B/32,
ViT-B/16, and ViT-L/14 (8-task accuracy, \%). Depth-only
heuristics include linear and multi-stage tiered splits
(2/3/6/12). Composite uses EER/CCC-derived
adaptive scaling and provides consistent improvements.
}
\label{tab:scaling_strategies_backbones}
\end{table*}

\subsection{FusionBench Configuration and Evaluation Protocol}
All experiments utilize the \textbf{FusionBench} platform to evaluate the LARV across three \textbf{CLIP Vision Transformers} (ViT-B/32, ViT-B/16, ViT-L/14). Merging is assessed in three multi-task scenarios: the canonical 8-task, the challenging 14-task, and the high-interference 20-task classification suites.

\textbf{Primary Metric.} Performance is measured by \textbf{Top-1 Accuracy} (in \%). Results for the multi-task suites are reported as the \textbf{arithmetic mean} of the Top-1 accuracies across all $K$ tasks within that setting.

\textbf{Robustness and Generalization.}We evaluate robustness under distribution shift following the ImageNet-C protocol, testing seven common corruption types (Motion, Impulse, Gaussian, Pixelate, Spatter, Contrast, and JPEG) across four datasets.  
The main paper reports only the average corruption performance for clarity (Sec.~\ref{sec:corruption}), while the complete per-corruption results and detailed analysis for all datasets are included in Appendix~\ref{sec:full_corruption}.  
We additionally assess generalization by evaluating the merged models on two \textbf{unseen tasks} not used in constructing the task vectors.

\subsection{Hyperparameters and Implementation Details}
All baselines are run with the hyperparameter choices suggested in their original implementations, which we take as the default values in our evaluations.
\begin{itemize}
    \item \textbf{Task Arithmetic}: global coefficient $0.3$.
    
    \item \textbf{Iso-C}: the merge strength increases slightly with model size; we use factors of $1.30$, $1.40$, and $1.50$ for ViT-B/32, ViT-B/16, and ViT-L/14, respectively.
    
    \item \textbf{Iso-CTS}: scaling $1.5$ following the recommended configuration.
    
    \item \textbf{TIES-Merging}: coefficient $0.3$ as used in the original implementation.
    
    \item \textbf{TSV-M}: correction weight $\alpha = 1$, following the default
configuration used in the official FusionBench implementation.
\end{itemize}


\section{Ablation on Scaling Strategy: Beyond Simple Depth Priors}
\label{app:scaling}

\begin{table}[t]
\centering
\small
\setlength{\tabcolsep}{6pt}
\renewcommand{\arraystretch}{0.7}
\begin{tabular}{@{}ccc|cccc@{}}
\toprule

\multicolumn{3}{c|}{\textbf{Depth strategy}} &
\multirow{2}{*}{\textbf{TIES}} & \multirow{2}{*}{\textbf{ISO-C}} & \multirow{2}{*}{\textbf{ISO-CTS}} & \multirow{2}{*}{\textbf{TSV-M}} \\

$s_s$ & $s_m$ & $s_d$&&&\\
\midrule

\multicolumn{7}{c}{\textbf{Baseline}}\\
\cmidrule(lr){1-7}
1.0 & 1.0 & 1.0 & 71.9 & 82.5 & 82.7 & 83.1 \\
\midrule
\multicolumn{7}{c}{\textbf{Uniform}}\\
\cmidrule(lr){1-7}

0.5 & 0.5 & 0.5 & 64.6 & 74.6 & 74.8 & 77.6 \\
1.5 & 1.5 & 1.5 & 72.7 & 81.3 & 81.2 & 78.2 \\

\midrule
\multicolumn{7}{c}{\textbf{Freeze Shallow Layers}}\\
\cmidrule(lr){1-7}

0.0 & 0.0 & 0.5 & 55.8 & 61.5 & 61.8 & 63.5 \\
0.0 & 0.0 & 1.0 & 61.2 & 69.2 & 69.8 & 71.0 \\
0.0 & 0.0 & 1.5 & 65.2 & 73.9 & 74.3 & 74.3 \\
0.0 & 1.0 & 0.5 & 68.1 & 77.1 & 77.7 & 78.6 \\
0.0 & 1.0 & 1.0 & 72.5 & 81.8 & 82.5 & 82.4 \\
0.0 & 1.0 & 1.5 & 75.4 & 83.7 & 84.3 & 83.4 \\
0.0 & 1.5 & 0.5 & 70.7 & 78.5 & 79.4 & 78.4 \\
0.0 & 1.5 & 1.0 & 74.7 & 82.5 & 83.2 & 81.3 \\
0.0 & 1.5 & 1.5 & 77.0 & 84.0 & 84.6 & 81.8 \\

\midrule
\multicolumn{7}{c}{\textbf{Freeze Deep Layers}}\\
\cmidrule(lr){1-7}

0.0 & 0.5 & 0.0 & 56.2 & 62.1 & 62.4 & 63.3 \\
0.0 & 1.0 & 0.0 & 61.7 & 68.7 & 69.4 & 68.6 \\
0.0 & 1.5 & 0.0 & 65.2 & 71.2 & 72.2 & 69.8 \\
0.5 & 0.5 & 0.0 & 58.0 & 64.5 & 64.4 & 65.9 \\
0.5 & 1.0 & 0.0 & 63.1 & 70.3 & 70.6 & 70.8 \\
0.5 & 1.5 & 0.0 & 66.2 & 72.4 & 73.1 & 71.8 \\
1.5 & 0.5 & 0.0 & 53.7 & 61.8 & 61.1 & 60.3 \\
1.5 & 1.0 & 0.0 & 58.6 & 67.0 & 66.7 & 65.6 \\
1.5 & 1.5 & 0.0 & 61.6 & 68.4 & 68.5 & 66.4 \\

\bottomrule
\end{tabular}
\caption{
Depth-only structural baselines across backbones on ViT-B/32.
We use the same depth-scaling notation as Fig.~\ref{fig:tiered_sensitivity},
where $s_s$, $s_m$, and $s_d$ denote the shallow-, middle-, and deep-layer
scales, respectively. The first three columns list the specific
$(s_s,s_m,s_d)$ configuration, while the remaining columns report accuracy
(no \%) for four merge rules. \emph{Uniform} sets all three tiers to the same
scale; \emph{Freeze Shallow} fixes $s_s=0$ and varies $(s_m,s_d)$; and
\emph{Freeze Deep} fixes $s_d=0$ and varies $(s_s,s_m)$.
}
\label{tab:freeze_depth_backbones}
\end{table}

\begin{figure*}[t]
    \centering
    \includegraphics[width=\linewidth]{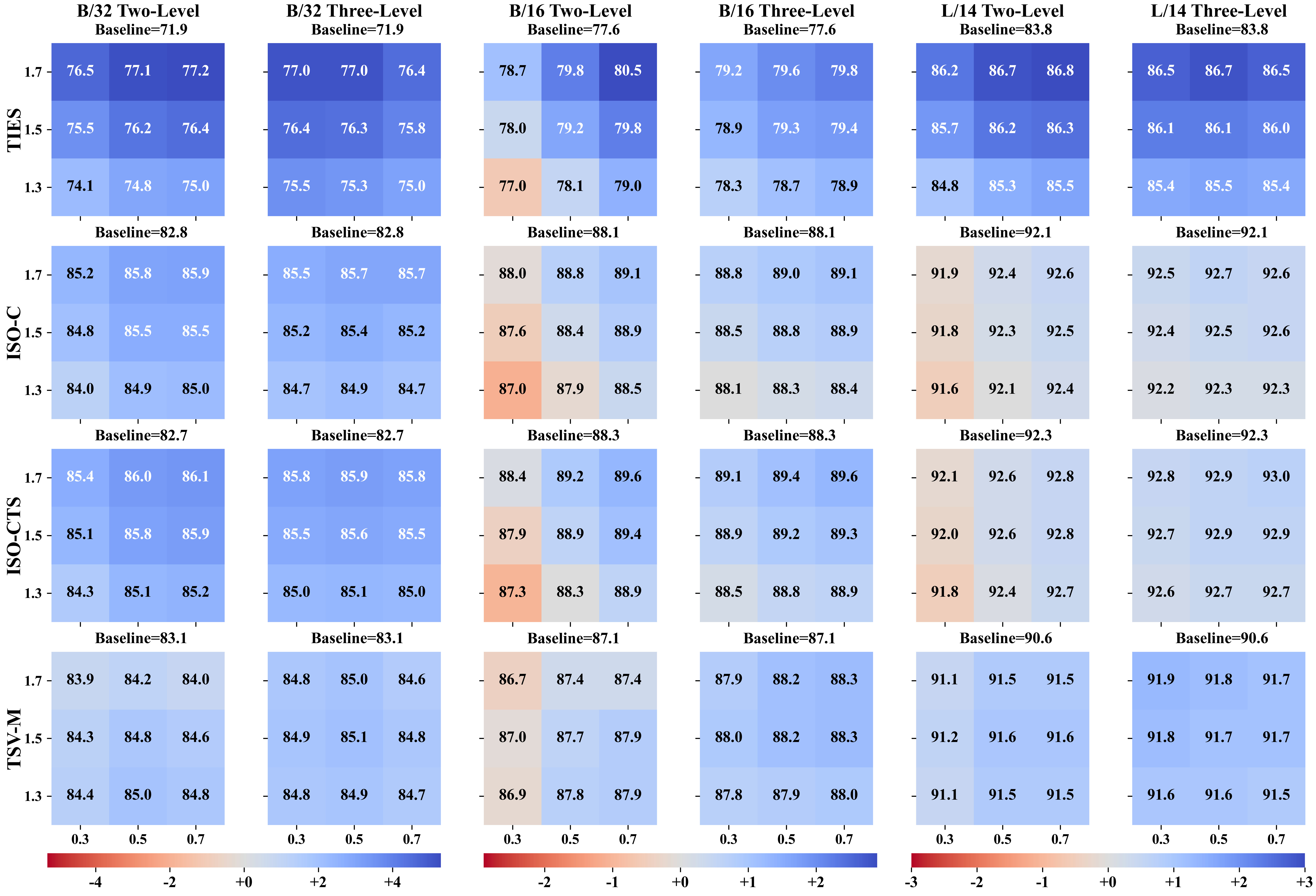}
\caption{
\textbf{Sensitivity of the tiered scaling scheme across eight vision tasks.}
Each heatmap reports the $\Delta$ accuracy (LARV $-$ Base, in percentage points)
for a particular choice of tiered scaling coefficients.
Positive values (\textcolor{red}{red}) indicate improvements, negative values (\textcolor{blue}{blue}) indicate decreases, and the colormap is centered at zero so that color intensity reflects the magnitude of deviation from the baseline.
The results show that moderate adjustments to middle or deep layers consistently provide the largest gains, highlighting the robustness of the tiered design across architectures.
}

\vspace{3mm}
    \label{fig:tiered_sensitivity}
\end{figure*}

\subsection{Comprehensive Comparison to Depth-Only Heuristics}
\label{app:scaling_strategies}

In section \ref{sec:composite-gate} we discussed the tier-gate strategy that categorizes the layers and assign distinct scaling factors by the depth prior. To contextualize the proposed LARV scaling rule, we compare it against several depth-only heuristics that assign fixed scale values based solely on layer index. Throughout this section, we use the unified notation 
$\mathbf{s_s}$ (shallow-layer scale), 
$\mathbf{s_m}$ (middle-layer scale), and 
$\mathbf{s_d}$ (deep-layer scale) 
to describe tiered schedules.  
These schedules do not rely on weight-derived signals such as $e_\ell$ or $c_\ell$, and therefore represent purely structural depth priors.

\begin{figure*}[t]
    \centering
    \includegraphics[width=\linewidth]{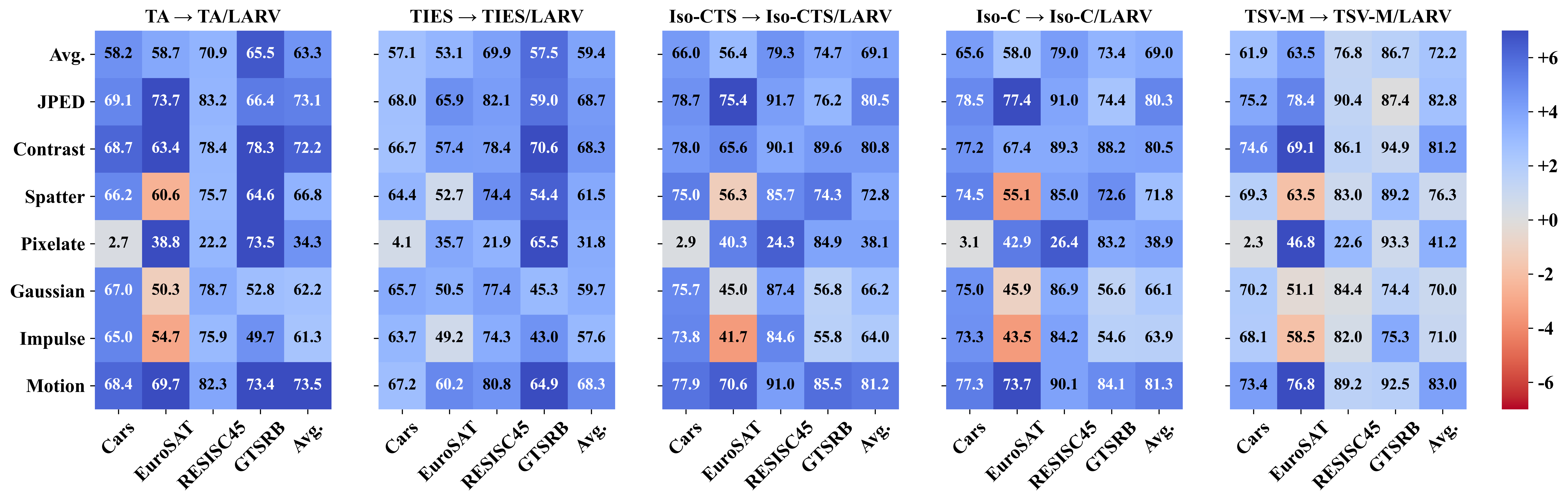}
    \caption{
\textbf{Corruption-wise accuracy gains introduced by LARV.}
Each heatmap visualizes the \emph{$\Delta$ accuracy} (LARV $-$ Base, in percentage points) across eight corruption types (Motion, Impulse, Gaussian, Pixelate, Spatter, Contrast, JPEG, and Avg.) evaluated on four datasets (Cars, EuroSAT, RESISC45, GTSRB) and their averaged performance. 
A positive delta indicates that LARV improves robustness, while a negative delta reflects a decrease in accuracy.
The color map is centered at zero: \textcolor{red}{red} denotes negative changes, \textcolor{blue}{blue} denotes positive improvements, and darker shades indicate larger magnitude.}

\vspace{3mm}
    \label{fig:corruption_full}
\end{figure*}

\paragraph{Depth-only heuristics.}
We evaluate the following baseline strategies:
\begin{itemize}
    \item \textbf{Uniform}: a single global scaling coefficient.
    \item \textbf{Tier2}: two-stage schedule with 
    $s_s = 0.5$ for the shallow half and 
    $s_d = 1.5$ for the deep half (with $s_m = 1$ implicitly).
    \item \textbf{Tier3}: three-stage monotonic schedule  
    $(s_s, s_m, s_d) = (0.5, 1.0, 1.5)$.
    \item \textbf{Tier6}: six-stage coarse depth schedule  
    $(0.5, 0.7, 0.9, 1.1, 1.3, 1.5)$.
    \item \textbf{Tier12}: fine-grained twelve-stage schedule  
    increasing linearly from $0.5$ to $1.5$.
    \item \textbf{Linear}: continuous interpolation from  
    $s_s = 0.5$ (shallow) to $s_d = 1.5$ (deep).
\end{itemize}

For backbones with different
depths (e.g., ViT-B/32, ViT-B/16, ViT-L/14), this schedule
is mapped proportionally across the $L$ layers so that the
relative progression remains consistent. For comparison, we also include our full LARV scaling rule, which integrates the
weight-derived EER and CCC signals into a composite score followed by a bounded
nonlinearity. This strategy adapts the scaling value for each layer based on its
learned behavior rather than its depth alone, and therefore departs fundamentally
from depth-index heuristics.

Table~\ref{tab:scaling_strategies_backbones} summarizes the 8-task average accuracy for four merging frameworks (TIES, ISO-C, ISO-CTS, TSV-M) on three ViT backbones (B/32, B/16, L/14).
The results show that depth-only schedules—whether using uniform scaling, linear schemes, or multi-stage tiered splits—offer only mild and often inconsistent benefits when the backbone architecture changes.
In comparison, LARV’s EER/CCC-guided layer-wise scaling improves all four merging methods on every backbone, suggesting that data-driven layer behavior provides a more stable and effective signal than simple depth-based heuristics.

\subsection{Structural Analysis via Layer Freezing: Validating Depth Heterogeneity}
\label{app:freeze}

To further analyze how different depth regions contribute to successful
model merging, we conduct a series of depth-only structural ablations on
ViT-B/32. These experiments freeze either the shallow layers or the deep
layers, isolating how performance changes when specific portions of the
network are prevented from adapting. The goal is not to propose new
merging strategies, but to provide controlled evidence for the
depth-dependent behavior observed throughout the main paper.

\paragraph{Experimental setup.}
We evaluate three families of depth configurations:
\begin{itemize}
    \item \textbf{Uniform scaling}: all layers share the same scaling factor,
          such as $(0.5,0.5,0.5)$, $(1.0,1.0,1.0)$, and $(1.5,1.5,1.5)$.
    \item \textbf{Freeze Shallow}: the shallow blocks are fixed
          (e.g., $(0.0,\cdot,\cdot)$), while the middle and deep layers
          remain scalable.
    \item \textbf{Freeze Deep}: the deep blocks are fixed
          (e.g., $(\cdot,\cdot,0.0)$), while earlier layers remain scalable.
\end{itemize}

These configurations represent deliberately extreme manipulations of depth. Unlike LARV's three-stage scheme, which modulates shallow, middle, and deep
layers smoothly, the freeze-based settings intentionally disrupt specific
regions to reveal where useful semantic information resides.

\paragraph{Interpreting Table~\ref{tab:freeze_depth_backbones}.}
Across all four merge rules, a consistent pattern emerges:
\begin{itemize}
    \item \textbf{Freezing deep layers is the most harmful.}
          For every deep-frozen configuration, accuracy drops sharply
          (e.g., TIES: $56.2 \rightarrow 61.6$, ISO-C: $62.1 \rightarrow 68.4$),
          reflecting the loss of semantic alignment normally provided by the
          deepest blocks.
    \item \textbf{Freezing shallow layers results in much milder degradation.}
          Under shallow-freezing, performance remains relatively high
          (e.g., ISO-C reaches $84.0$ and ISO-CTS reaches $84.6$),
          indicating that early layers largely encode task-specific or noisy
          variations that are less critical for successful merging.
    \item \textbf{Both extremes underperform uniform scaling.}
          Even the best freeze-based results fall short of uniform
          configurations such as $(1.0,1.0,1.0)$, highlighting that neither
          the shallow nor the deep region can be removed from adaptation
          without cost.
\end{itemize}

These structural ablations confirm the core motivation behind the
three-stage depth-aware scaling strategy. Shallow layers tend to be noisy
and should not dominate the merge; deep layers carry the strongest
task-invariant semantics and must be scaled more aggressively; and the
middle region acts as a stabilizing transition. By smoothly adjusting
scaling across depth rather than freezing or treating all layers equally,
LARV preserves the strengths of each depth region and produces more robust
merged models.

\subsection{Tiered Scaling Robustness and Sensitivity Analysis}
\label{app:two-vs-three}
Figure~\ref{fig:tiered_sensitivity} provides a detailed sensitivity analysis of the two-level and three-level tiered scaling schemes across eight vision tasks and three ViT backbones (B/32, B/16, L/14). Several consistent and architecture-invariant patterns emerge.

First, \textbf{deep-layer amplification ($s_d$) is universally beneficial}. For all merging rules (TIES, Iso-C, Iso-CTS, TSV-M), increasing $s_d$ from $1.0$ to $1.5$ yields monotonic gains, particularly on B/32 and B/16. This trend aligns with our diagnostic metrics: deeper blocks exhibit higher information richness ($e_\ell$) and lower conflict ($c_\ell$), so amplifying their contributions reliably strengthens semantic alignment across tasks.

Second, \textbf{shallow-layer shrinkage ($s_s$) consistently prevents degradation}. Across all backbones, settings with $s_s=0.5$ outperform or match $s_s=1.0$, while aggressive amplification ($s_s=1.5$) frequently harms performance—especially for TIES and ISO-C on B/32. This supports our central hypothesis that early layers contain noisy or task-specific variations that should not be emphasized during merging.

Third, \textbf{three-level schemes offer smoother, more stable behavior}. While two-level schedules already outperform uniform scaling, the three-level variant reduces performance variance by adding a flexible middle tier ($s_m$). This middle bucket prevents over-shrinkage or over-amplification, leading to more robust cross-rule improvements. The effect is particularly visible on Iso-CTS and TSV-M, where most $(s_s,s_m,s_d)$ configurations form a flat plateau of strong results around $(0.5,1.0,1.5)$.

Finally, \textbf{scaling trends persist across backbones and merging rules}, demonstrating that layer-wise heterogeneity is a structural, architecture-agnostic property of ViTs rather than an artifact of a specific rule or dataset. Larger backbones (e.g., ViT-L/14) show slightly reduced sensitivity overall, suggesting that increased capacity mitigates interference but still benefits from depth-aware scaling.

Overall, the sensitivity maps reinforce the design of LARV's tiered scaling: suppress noisy shallow updates, stabilize mid-depth transitions, and amplify deep semantic structure. This explains why LARV remains effective across merge rules and backbones despite using a fixed, data-free policy.

\section{Full Results on Robustness and Generalization}

\subsection{Detailed Results on Corruption Tasks}
\label{sec:full_corruption}
In the main paper, we use radar plots to summarize each method’s robustness trend across corruption types, providing a high-level comparison of the baseline mergers and their LARV-enhanced variants. While these visualizations reveal the overall shape of performance differences, they do not expose the fine-grained behavior of each dataset under specific corruption settings.

To provide a complete picture, this section reports the full corruption-wise performance details. Instead of repeating raw accuracies, we visualize the \emph{$\Delta$ accuracy} (LARV $-$ Base, in percentage points) for all eight corruption types across the four datasets and their averaged performance. This representation makes the effect of LARV immediately interpretable: positive $\Delta$ values indicate robustness improvements, negative values indicate decreases, and the diverging colormap centers zero to highlight the direction and magnitude of change. 

These detailed heatmaps reveal several patterns that complement the aggregated observations in the main text. LARV consistently improves robustness across most corruptions and methods, but the gains are not uniform. High-frequency distortions such as Pixelate, Spatter, and Contrast show particularly large improvements, suggesting that LARV stabilizes deeper transformer blocks that are especially sensitive to such perturbations. Even when baseline performance is strong, LARV provides additional robustness headroom without degrading accuracy on simpler corruptions.

Overall, these detailed results corroborate the conclusions drawn from the radar plots: LARV yields systematic, architecture-agnostic robustness gains and provides a more reliable corruption response profile across merging methods.

\begin{table*}[t]
\centering
\small
\setlength{\tabcolsep}{4pt}
\renewcommand{\arraystretch}{0.92}

\resizebox{\linewidth}{!}{
\begin{tabular}{@{}l|c|*{15}{c}@{}}
\toprule
\textbf{Method} & \textbf{LARV} &
\textbf{SUN397} & \textbf{Cars} & \textbf{RESISC45} &
\textbf{EuroSAT} & \textbf{SVHN} & \textbf{GTSRB} & \textbf{MNIST} &
\textbf{DTD} & \textbf{Flowers} & \textbf{PCAM} & \textbf{FER2013} &
\textbf{Pets} & \textbf{STL10} & \textbf{CIFAR100} & \textbf{Avg.} \\
\midrule
\multicolumn{17}{c}{\textbf{ViT-B/32}} \\
\cmidrule(lr){2-17}

Fine-tuned & - &
82.8 & 92.8 & 97.4 & 99.1 & 97.9 & 99.2 &
99.8 & 85.5 & 97.7 & 91.1 & 75.9 & 95.8 &
99.2 & 93.0 & 93.4\phantom{0} \\

Simple Avg. & - &
64.8 & 60.4 & 67.1 & 67.0 & 50.7 & 45.6 &
76.6 & 46.9 & 67.4 & 65.2 & 51.6 & 84.2 &
97.2 & 70.4 & 65.4\phantom{0} \\

\rowcolor{black!5} & - &
41.8 & 33.2 & 47.3 & 55.4 & 46.5 & 48.4 &
88.7 & 37.0 & 38.6 & 64.1 & 46.1 & 65.9 &
84.6 & 41.7 & 52.8\phantom{0} \\

\rowcolor{black!5}\multirow{-2}{*}{Task Arithmetic} & \checkmark &
51.1 & 44.4 & 60.0 & 82.6 & 68.3 & 65.9 &
96.3 & 46.1 & 50.0 & 70.1 & 57.3 & 76.8 &
93.5 & 59.2 & 65.8\gainu{$\uparrow$} \\

& - &
69.5 & 62.6 & 79.8 & 83.4 & 78.4 & 84.1 &
97.0 & 61.2 & 69.3 & 83.6 & 65.6 & 87.1 &
96.8 & 74.3 & 78.1\phantom{0} \\

\multirow{-2}{*}{ISO-C} & \checkmark &
72.1 & 63.0 & 87.4 & 91.2 & 77.7 & 87.7 &
97.7 & 68.9 & 73.5 & 82.2 & 68.4 & 88.5 &
97.8 & 80.2 & 81.2\gainu{$\uparrow$} \\

\rowcolor{black!5}& - &
69.7 & 68.0 & 81.6 & 85.3 & 78.1 & 85.7 &
97.5 & 62.9 & 73.5 & \textbf{84.6} & 67.1 & 87.0 &
97.0 & 74.2 & 79.4\phantom{0} \\

\rowcolor{black!5}\multirow{-2}{*}{ISO-CTS} & \checkmark &
\textbf{72.2} & \textbf{69.5} & \textbf{89.4} &
91.4 & 77.0 & 88.7 &
97.8 & \textbf{71.4} & \textbf{79.8} &
81.4 & \textbf{69.1} & 87.6 &
\textbf{97.9} & \textbf{80.7} & \textbf{82.4}\gainu{$\uparrow$} \\

& - &
62.2 & 54.6 & 65.3 & 63.0 & 65.7 & 63.9 &
92.6 & 49.9 & 58.2 & 77.1 & 54.9 & 81.4 &
94.8 & 62.4 & 67.6\phantom{0} \\

\multirow{-2}{*}{TIES} & \checkmark &
63.3 & 56.7 & 74.4 & 83.3 & 75.9 & 73.9 &
96.3 & 54.3 & 61.7 & 76.6 & 61.9 & 83.0 &
97.0 & 72.1 & 73.6\gainu{$\uparrow$} \\

\rowcolor{black!5}& - &
66.5 & 62.6 & 80.7 & 91.3 & \textbf{86.7} & 89.3 &
98.6 & 60.9 & 66.2 & 81.3 & 65.3 & 87.4 &
96.7 & 69.5 & 78.8\phantom{0} \\

\rowcolor{black!5}\multirow{-2}{*}{TSV-M} & \checkmark &
67.4 & 63.3 & 85.8 & \textbf{94.7} & 86.0 & \textbf{90.8} &
\textbf{98.7} & 67.8 & 72.2 & 80.5 & 68.2 & \textbf{89.0} &
97.8 & 74.9 & 81.2\gainu{$\uparrow$} \\
\midrule
\multicolumn{17}{c}{\textbf{ViT-B/16}} \\
\cmidrule(lr){2-17}

Fine-tuned & - &
78.9 & 85.9 & 96.6 & 99.0 & 97.6 & 99.0 &
99.7 & 82.3 & 94.9 & 90.6 & 72.8 & 94.5 &
98.2 & 88.8 & 91.3\phantom{0} \\

Simple Avg. & - &
67.5 & 65.9 & 71.5 & 71.1 & 64.6 & 54.1 &
82.6 & 47.2 & 72.5 & 63.2 & 54.1 & 90.4 &
98.3 & 73.4 & 69.7\phantom{0} \\

\rowcolor{black!5}& - &
52.5 & 40.8 & 52.0 & 50.5 & 59.4 & 50.7 &
90.1 & 37.1 & 49.8 & 82.4 & 55.6 & 87.2 &
92.8 & 50.0 & 60.8\phantom{0} \\

\rowcolor{black!5}\multirow{-2}{*}{Task Arithmetic} & \checkmark &
62.4 & 49.0 & 73.8 & 79.9 & 77.2 & 71.5 &
97.2 & 45.7 & 63.6 & 82.9 &
65.1 & 92.4 &
96.7 & 68.8 & 73.3\gainu{$\uparrow$} \\

& - &
72.8 & 66.2 & 88.2 & 91.8 & 86.7 & 87.7 &
98.2 & 60.1 & 81.0 & 82.6 & 67.0 & 93.9 &
98.4 & 79.5 & 82.4\phantom{0} \\

\multirow{-2}{*}{ISO-C} & \checkmark &
74.4 & 63.0 & 91.3 & 95.7 & 86.0 & 89.3 &
98.4 & 68.7 & 86.1 & 82.4 & 70.1 & 94.0 &
98.9 & \textbf{82.7} & 84.3\gainu{$\uparrow$} \\

\rowcolor{black!5}& - &
73.5 & \textbf{74.1} & 90.1 & 94.3 & \textbf{89.0} & 90.4 &
98.3 & 65.3 & 85.5 & 81.2 & 68.6 & 93.8 &
98.5 & 80.0 & 84.5\phantom{0} \\

\rowcolor{black!5}\multirow{-2}{*}{ISO-CTS} & \checkmark &
\textbf{74.9} & 72.9 & \textbf{92.5} & \textbf{96.2 }& 87.3 & \textbf{90.9} &
98.5 & \textbf{73.7} & \textbf{91.0} & 82.1 & \textbf{70.9} & 93.7 &
\textbf{98.9} & 82.5 & \textbf{86.1}\gainu{$\uparrow$} \\

& - &
66.9 & 60.3 & 71.9 & 69.8 & 71.4 & 64.4 &
93.6 & 49.0 & 64.8 & 73.1 & 59.6 & 90.8 &
97.6 & 67.9 & 71.5\phantom{0} \\

\multirow{-2}{*}{TIES} & \checkmark &
69.4 & 60.7 & 82.1 & 84.9 & 78.4 & 77.1 &
96.9 & 54.1 & 70.6 & 71.8 & 67.2 & 93.3 &
98.3 & 77.5 & 77.3\gainu{$\uparrow$} \\

\rowcolor{black!5}& - &
70.1 & 66.9 & 85.6 & 92.6 & 88.7 & 88.4 &
98.7 & 59.3 & 77.5 & \textbf{84.4} & 69.7 & 93.6 &
98.0 & 76.3 & 82.1\phantom{0} \\

\rowcolor{black!5}\multirow{-2}{*}{TSV-M} & \checkmark &
71.8 & 67.1 & 89.7 & 96.2 & 88.4 & 90.5 &
\textbf{98.9} & 66.4 & 83.9 & 82.8 & 70.8 & \textbf{94.1} &
98.5 & 80.0 & 84.2\gainu{$\uparrow$} \\

\midrule
\multicolumn{17}{c}{\textbf{ViT-L/14}} \\
\cmidrule(lr){2-17}

Fine-tuned & - &
74.9 & 78.5 & 95.1 & 99.1 & 97.3 & 98.9 &
99.6 & 79.7 & 88.6 & 88.0 & 71.6 & 92.5 &
97.6 & 88.4 & 89.3\phantom{0} \\

Simple Avg. & - &
71.2 & 79.0 & 78.7 & 80.4 & 71.3 & 64.6 &
94.3 & 58.7 & 81.9 & 74.2 & 54.8 & 94.6 &
99.3 & 82.4 & 77.5\phantom{0} \\

\rowcolor{black!5}& - &
60.6 & 53.2 & 48.1 & 53.0 & 50.1 & 54.2 &
93.0 & 41.6 & 59.6 & 75.8 & 53.9 & 89.3 &
94.2 & 57.2 & 63.1\phantom{0} \\

\rowcolor{black!5}\multirow{-2}{*}{Task Arithmetic} & \checkmark &
68.5 & 68.6 & 70.7 & 77.6 & 77.1 & 72.1 &
97.7 & 53.6 & 77.9 & 78.5 & 60.8 & 93.3 &
98.2 & 74.0 & 76.3\gainu{$\uparrow$} \\

& - &
78.9 & 89.1 & 93.8 & 94.5 & 91.8 & 94.8 &
98.8 & 74.4 & 96.7 & 85.8 & 70.9 & \textbf{96.6} &
99.6 & 87.7 & 89.5\phantom{0} \\

\multirow{-2}{*}{ISO-C} & \checkmark &
80.2 & 90.4 & 94.8 & 96.8 & 90.6 & 94.7 &
98.9 & 79.0 & 97.8 & 85.1 & 73.5 & 96.5 &
99.5 & \textbf{89.6} & 90.5\gainu{$\uparrow$} \\

\rowcolor{black!5}& - &
79.8 & 90.7 & 94.8 & \textbf{95.9} & 92.6 & 95.9 &
\textbf{99.0} & 78.0 & 97.4 & 84.1 & 72.8 & \textbf{96.6}&
99.6 & 88.4 & 90.4\phantom{0} \\

\rowcolor{black!5}\multirow{-2}{*}{ISO-CTS} & \checkmark &
\textbf{80.6} & \textbf{91.8} & \textbf{95.7} & \textbf{97.4} & 91.1 & 95.6 &
98.9 & \textbf{81.4} & \textbf{98.0} & 84.6 & \textbf{74.3} & 96.3 &
99.5 & \textbf{89.6} & \textbf{91.1}\gainu{$\uparrow$} \\

& - &
72.0 & 75.6 & 76.5 & 69.7 & 77.2 & 75.1 &
96.6 & 57.8 & 79.6 & 78.2 & 60.0 & 94.7 &
98.4 & 77.7 & 77.8\phantom{0} \\

\multirow{-2}{*}{TIES} & \checkmark &
74.3 & 80.7 & 84.9 & 89.1 & 84.5 & 83.0 &
98.2 & 63.6 & 87.2 & 83.1 & 64.7 & 95.4 &
99.1 & 84.5 & 83.7\gainu{$\uparrow$} \\

\rowcolor{black!5}& - &
76.0 & 86.7 & 91.1 & 94.2 & \textbf{93.2} &
93.8 & \textbf{99.0} & 69.5 & 95.1 & \textbf{86.0} & 69.8 &
96.3 & 99.2 & 85.2 & 88.2\phantom{0} \\

\rowcolor{black!5}\multirow{-2}{*}{TSV-M} & \checkmark &
77.8 & 89.1 & 93.0 & 97.3 &
92.8 & 94.4 &
\textbf{99.0} & 76.1 & 97.1 & 84.2 & 72.9 & 96.3 &
99.3 & 87.7 & 89.8\gainu{$\uparrow$} \\

\bottomrule
\end{tabular}}
\caption{
\textbf{Results on 14 classification tasks across five ViT backbones.}
For each backbone (ViT-B/32, ViT-B/16, ViT-L/14), we report the accuracy (\%) of seven merging methods, with and without LARV. 
Fine-tuned models are included as an upper bound. 
Best results per task (excluding Fine-tuned) are highlighted in bold.
This table complements the high-level robustness trends shown in the main paper by providing per-dataset accuracy details, demonstrating that LARV consistently improves or matches performance across diverse tasks and architectures.
}
\label{tab:14tasks_all}
\end{table*}

\subsection{Generalization Beyond Vision}
To assess modality generality, we evaluate LARV on NLP tasks using T5-LoRA adapters across eight language tasks. Without any re-tuning, LARV consistently improves all base merging rules, indicating that layer-wise interference patterns and weight-only diagnostics extend beyond vision transformers. Detailed results are provided in the Appendix.
\begin{table}[t]
\centering
\small
\setlength{\tabcolsep}{12pt}
\renewcommand{\arraystretch}{0.9}
\resizebox{0.95\linewidth}{!}{
\begin{tabular}{l|c|c|c|c|c|c}
\toprule
 & \textbf{LARV} & \textbf{TA} & \textbf{ISO-C} & \textbf{ISO-CTS} & \textbf{TIES} & \textbf{TSV-M} \\
\midrule
T5-LoRA & -          & 77.41\phantom{$\uparrow$} & 76.28\phantom{$\uparrow$} & 76.26\phantom{$\uparrow$} & 76.79\phantom{$\uparrow$} & 81.83\phantom{$\uparrow$} \\
T5-LoRA & \checkmark & 78.00$\uparrow$ & 76.44$\uparrow$ & 76.43$\uparrow$ & 77.52$\uparrow$ & 82.21$\uparrow$ \\
\bottomrule
\end{tabular}
}
\caption{Additional Experiments on 8 NLP Tasks.}
\end{table}

\subsection{Detailed Results on 14-Task and 20-Task Merges}

Tables~7 and~8 provide the complete accuracy results for the 14-task and 20-task merging benchmarks, complementing the aggregated performance curves shown in the main paper. While the main text focuses on overall improvements and backbone-level trends, the full tables reveal several deeper insights into how LARV behaves across tasks, architectures, and merge rules.

First, LARV produces improvements that are surprisingly stable across both the 14-task and 20-task settings, despite the substantial increase in task diversity. Moving from 14 to 20 tasks introduces broader variations in granularity (e.g., GTSRB vs.\ Food101), modality shift (e.g., PCAM and FER2013), and texture-heavy domains (e.g., DTD). Traditional merge baselines struggle in such heterogeneous settings because scaling errors propagate unevenly across layers; LARV counteracts this by rebalancing shallow, middle, and deep representations in a task-agnostic way, enabling consistent gains even when the underlying task mixture becomes more challenging.

Second, the tables highlight that LARV’s benefits are not uniform but follow an interpretable pattern aligned with the analysis in the main paper.
Fine-grained and structure-dependent datasets (Cars, Pet Images, Flowers102, STL10, CIFAR100) tend to benefit the most. These tasks rely heavily on spatially precise or high-frequency cues, which are sensitive to mismatched feature magnitudes after merging. LARV’s adaptive depthwise rescaling helps recover these signals, especially in deeper layers where semantic abstraction is formed. Conversely, tasks with extremely low intrinsic dimensionality (e.g., MNIST, KMNIST) exhibit smaller but still positive gains, consistent with the idea that early transformer blocks dominate their decision boundaries.

Third, cross-backbone consistency offers additional evidence that the LARV mechanism generalizes beyond any particular model scale. ViT-B/32 gains are the largest in relative terms, reflecting that coarse-grained early representations are particularly vulnerable to scaling mismatch. In contrast, ViT-L/14—whose deeper feature hierarchy already exhibits stable layer statistics—still benefits from LARV, but the improvements manifest more in robustness across tasks than in raw accuracy. This agrees with our findings in §4 that deeper backbones exhibit smaller variance in their layer-wise magnitudes, yet still profit from controlled rescaling at block level.

Finally, the detailed results also reveal systematic differences between merge rules. Methods with inherently stronger geometric alignment (ISO-C, ISO-CTS) tend to show more moderate but still consistent gains under LARV, whereas magnitude-sensitive mergers (TIES, TSV-M) show the largest absolute improvements. This pattern aligns with the theory that LARV corrects scaling distortions that accumulate during magnitude-based merging, especially in deeper layers. Importantly, LARV never degrades performance in a systematic way, demonstrating that its veneer formulation serves as a reliable drop-in enhancement to a wide range of merging templates.

Overall, the complete results in Tables~7 and~8 substantiate the main claim of the paper: LARV improves multi-task model merging not simply through uniform rescaling, but through a principled depth-aware correction that stabilizes representations across a wide spectrum of datasets, architectures, and merging paradigms.

\begin{table*}[t]
\centering
\small
\setlength{\tabcolsep}{4pt}
\renewcommand{\arraystretch}{0.92}
\resizebox{\linewidth}{!}{
\begin{tabular}{@{}l|c|*{21}{c}@{}}
\toprule
\textbf{Method} & \textbf{LARV} &
\textbf{\#1} & \textbf{\#2} & \textbf{\#3} & \textbf{\#4} &
\textbf{\#5} & \textbf{\#6} & \textbf{\#7} & \textbf{\#8} &
\textbf{\#9} & \textbf{\#10} & \textbf{\#11} & \textbf{\#12} &
\textbf{\#13} & \textbf{\#14} & \textbf{\#15} & \textbf{\#16} &
\textbf{\#17} & \textbf{\#18} & \textbf{\#19} & \textbf{\#20} &
\textbf{Avg.} \\
\midrule

\multicolumn{23}{c}{\textbf{ViT-B/32}} \\
\cmidrule(lr){2-23}

Fine-tuned & - &
74.9 & 78.5 & 95.1 & 99.1 & 97.3 & 98.9 &
99.6 & 79.7 & 88.6 & 88.0 & 71.6 & 92.5 &
97.6 & 88.4 & 97.6 & 88.4 & 94.8 & 95.6 &
98.2 & 71.3 & 89.8\phantom{0} \\

Simple Avg. & - &
64.2 & 59.6 & 64.8 & 60.9 & 47.3 & 43.1 &
71.8 & 46.4 & 66.5 & 63.9 & 50.2 & 84.1 &
97.0 & 69.8 & 92.7 & 80.4 & 71.3 & 15.0 &
11.5 & 61.8 & 61.1\phantom{0} \\

\rowcolor{black!5}& - &
20.4 & 12.2 & 25.6 & 25.6 & 30.9 & 29.8 &
78.0 & 22.3 & 21.1 & 53.2 & 34.3 & 42.4 &
71.0 & 29.5 & 64.1 & 15.1 & 67.0 & 17.0 &
15.4 & 51.2 & 36.3\phantom{0} \\

\rowcolor{black!5}\multirow{-2}{*}{Task Arithmetic} & \checkmark &
34.6 & 23.0 & 40.6 & 55.1 & 57.3 & 49.7 &
94.5 & 33.8 & 35.1 & 60.5 & 49.6 & 61.6 &
88.2 & 45.0 & 83.3 & 32.6 & 76.2 & 24.4 &
22.7 & 51.2 & 51.0\gainu{$\uparrow$} \\

& - &
57.1 & 33.6 & 61.5 & 74.4 & 80.6 & 84.1 &
98.0 & 51.1 & 47.6 & 70.3 & 59.3 & 75.1 &
93.1 & 65.1 & 91.3 & 54.3 & 83.5 & 42.4 &
39.5 & 68.9 & 66.5\phantom{0} \\

\multirow{-2}{*}{ISO-C} & \checkmark &
61.6 & 28.7 & 79.3 & 89.3 & 80.9 & 89.5 &
98.3 & 62.0 & 55.9 & 77.8 & 66.0 & 75.1 &
95.7 & 73.3 & 93.9 & 71.2 & 86.8 & 50.9 &
55.8 & 71.1 & 73.2\gainu{$\uparrow$} \\

\rowcolor{black!5}& - &
57.0 & 47.2 & 64.7 & 75.7 & 81.6 & 86.7 &
98.0 & 54.4 & 57.3 & 78.9 & 62.0 & 77.5 &
94.3 & 67.2 & 92.4 & 58.0 & 84.8 & 47.7 &
50.9 & 68.1 & 70.2\phantom{0} \\

\rowcolor{black!5}\multirow{-2}{*}{ISO-CTS} & \checkmark &
62.7 & 44.4 & 84.2 & 90.2 & 81.6 & \textbf{90.7} &
\textbf{98.4} & \textbf{66.0} & 66.8 & \textbf{81.0} & 66.3 &
78.1 & 96.6 & \textbf{75.4} & \textbf{94.6} & 74.8 &
87.3 & \textbf{54.8} & 66.4 & 70.8 & 76.6\gainu{$\uparrow$} \\

& - &
51.0 & 36.2 & 47.8 & 45.1 & 58.2 & 57.7 &
92.1 & 40.6 & 44.8 & 66.9 & 47.3 & 73.1 &
89.9 & 51.3 & 86.3 & 50.1 & 76.5 & 21.0 &
19.7 & 55.9 & 55.6\phantom{0} \\

\multirow{-2}{*}{TIES} & \checkmark &
53.8 & 35.1 & 62.3 & 70.7 & 72.3 & 69.8 &
97.0 & 47.7 & 49.7 & 67.6 & 54.5 & 74.6 &
95.1 & 62.7 & 91.5 & 63.2 & 79.4 & 29.1 &
23.9 & 53.3 & 62.7\gainu{$\uparrow$} \\

\rowcolor{black!5}& - &
64.3 & \textbf{53.2} & 74.0 & 87.0 & 81.9 & 84.9 &
97.9 & 58.5 & 61.7 & 79.4 & 63.8 & 84.9 &
96.0 & 67.6 & 93.3 & 73.4 & 84.7 & 38.7 &
49.1 & 70.2 & 73.2\phantom{0} \\

\rowcolor{black!5}\multirow{-2}{*}{TSV-M} & \checkmark &
\textbf{65.3} & 50.9 & \textbf{81.5} & \textbf{93.5} &
\textbf{83.4} & 88.5 &
98.3 & 65.0 & \textbf{68.5} & 79.3 &\textbf{ 67.6} & \textbf{85.1} &
\textbf{97.3} & 72.4 & \textbf{94.6} & \textbf{79.3} &
\textbf{87.9} & 47.5 & \textbf{68.6} &
\textbf{71.4} & \textbf{77.3}\gainu{$\uparrow$} \\

\midrule
\multicolumn{23}{c}{\textbf{ViT-B/16}} \\
\cmidrule(lr){2-23}

Fine-tuned & - &
78.9 & 85.9 & 96.6 & 99.0 & 97.6 & 99.0 &
99.7 & 82.3 & 94.9 & 90.6 & 72.8 & 94.5 &
98.2 & 88.8 & 98.3 & 91.9 & 94.5 & 95.3 &
98.1 & 75.7 & 91.6\phantom{0} \\

Simple Avg. & - &
67.1 & \textbf{64.6} & 69.3 & 63.4 & 62.4 & 52.0 &
80.7 & 46.6 & 71.8 & 63.1 & 50.9 & 89.6 &
98.0 & 72.9 & 94.2 & 85.9 & 73.3 & 15.6 &
12.4 & 62.5 & 64.8\phantom{0} \\

\rowcolor{black!5}
& - &
26.0 & 12.3 & 21.9 & 26.3 & 35.0 & 30.4 &
69.4 & 23.8 & 22.3 & 69.8 & 38.4 & 64.8 &
80.4 & 31.3 & 64.9 & 19.4 & 64.0 & 17.8 &
24.9 & 54.8 & 39.9\phantom{0} \\

\rowcolor{black!5}
\multirow{-2}{*}{Task Arithmetic} & \checkmark &
48.8 & 23.0 & 45.8 & 49.1 & 62.7 & 48.1 &
93.2 & 34.1 & 46.7 & 82.2 & 57.5 & 85.4 &
93.4 & 54.8 & 88.6 & 57.0 & 83.2 & 36.1 &
43.8 & 53.2 & 59.3\gainu{$\uparrow$} \\

& - &
64.4 & 28.7 & 76.9 & 80.8 & 84.8 & 83.8 &
97.7 & 50.1 & 67.1 & 75.0 & 66.2 & 91.4 &
96.8 & 72.5 & 94.9 & 76.9 & 87.0 & 61.4 &
63.7 & 71.9 & 74.6\phantom{0} \\

\multirow{-2}{*}{ISO-C} & \checkmark &
65.3 & 17.8 & 86.7 & 93.5 & 86.2 & 88.2 &
98.5 & 60.8 & 74.1 & 78.2 & 69.9 & 91.6 &
98.2 & 77.7 & 96.1 & 85.7 & 88.9 & \textbf{75.6} &
71.3 & 71.6 & 78.8\gainu{$\uparrow$} \\

\rowcolor{black!5}& - &
66.9 & 45.9 & 82.7 & 86.9 & \textbf{88.2} & 89.2 &
98.6 & 57.6 & 76.1 & 70.5 & 67.9 & 91.8 &
97.5 & 74.8 & 95.2 & 81.7 & 88.0 & 62.6 &
66.6 & 73.9 & 78.1\phantom{0} \\

\rowcolor{black!5}
\multirow{-2}{*}{ISO-CTS} & \checkmark &
66.6 & 35.4 & \textbf{89.2} & 94.4 & 88.0 & \textbf{90.8} &
\textbf{98.8} & \textbf{67.8} & \textbf{82.6} & 76.0 & 70.4 & 92.0 &
98.2 & \textbf{78.3} & 96.1 & 87.4 & 88.8 & 72.3 &
72.3 & \textbf{75.2} & 81.0\gainu{$\uparrow$} \\

& - &
57.7 & 39.0 & 49.7 & 46.0 & 65.3 & 48.8 &
90.6 & 41.8 & 49.2 & 75.3 & 53.0 & 87.1 &
94.7 & 56.3 & 89.6 & 65.9 & 79.5 & 28.4 &
36.4 & 58.4 & 60.6\phantom{0} \\

\multirow{-2}{*}{TIES} & \checkmark &
63.1 & 35.6 & 71.8 & 71.6 & 75.5 & 63.2 &
96.2 & 47.6 & 60.0 & 77.0 & 63.3 & 89.8 &
97.5 & 69.4 & 94.7 & 79.7 & 86.4 & 51.0 &
46.2 & 56.0 & 69.8\gainu{$\uparrow$} \\

\rowcolor{black!5}& - &
67.7 & 52.5 & 78.4 & 86.5 & 83.7 & 81.3 &
97.2 & 54.5 & 72.9 & \textbf{85.5} & 67.3 & \textbf{92.9} &
97.5 & 72.8 & 95.1 & 83.8 & 88.5 & 62.7 &
70.8 & 74.5 & 78.3\phantom{0} \\

\rowcolor{black!5}\multirow{-2}{*}{TSV-M} & \checkmark &
\textbf{69.9} & 50.5 & 86.5 & \textbf{95.0} & 85.3 & 85.5 &
98.4 & 63.2 & 81.1 & 82.9 & \textbf{70.5} & \textbf{92.9} &
\textbf{98.4} & 77.0 & \textbf{96.2} & \textbf{88.3} & \textbf{89.4} & \textbf{75.6} &
\textbf{74.4} & 73.6 & \textbf{81.7}\gainu{$\uparrow$} \\

\midrule
\multicolumn{23}{c}{\textbf{ViT-L/14}} \\
\cmidrule(lr){2-23}

Fine-tuned & - &
82.8 & 92.8 & 97.4 & 99.1 & 97.9 & 99.2 &
99.8 & 85.5 & 97.7 & 91.1 & 75.9 & 95.8 &
99.2 & 93.0 & 99.1 & 94.8 & 95.3 & 95.4 &
98.3 & 80.5 & 93.5\phantom{0} \\

Simple Avg. & - &
70.7 & 77.7 & 76.4 & 75.3 & 69.5 & 62.1 &
93.7 & 57.7 & 80.8 & 73.6 & 52.7 & 94.2 &
99.2 & 81.7 & 97.0 & 90.7 & 77.4 & 16.1 &
10.4 & 66.1 & 71.1\phantom{0} \\

\rowcolor{black!5}
& - &
22.9 & 14.3 & 18.0 & 30.7 & 19.0 & 27.7 &
77.8 & 22.9 & 22.0 & 50.0 & 42.2 & 56.6 &
76.2 & 25.3 & 64.6 & 16.6 & 60.9 & 12.0 &
10.0 & 49.9 & 36.0\phantom{0} \\

\rowcolor{black!5}
\multirow{-2}{*}{Task Arithmetic} & \checkmark &
48.6 & 37.0 & 34.2 & 47.9 & 63.1 & 47.8 &
95.9 & 36.6 & 54.4 & 52.6 & 55.2 & 84.4 &
94.3 & 51.8 & 90.5 & 45.3 & 78.9 & 20.9 &
10.1 & 49.9 & 55.0\gainu{$\uparrow$} \\

& - &
77.9 & 80.3 & 91.5 & 91.3 & 89.4 & 92.9 &
98.8 & 71.1 & 95.9 & 85.3 & 69.2 & 96.4 &
99.3 & 86.0 & 98.0 & 92.1 & 90.6 & 46.5 &
40.2 & 76.3 & 83.4\phantom{0} \\

\multirow{-2}{*}{ISO-C} & \checkmark &
78.9 & 80.1 & 93.5 & 95.3 & 90.0 & 94.0 &
98.9 & 76.5 & 97.2 & 84.6 & 72.6 & 96.1 &
\textbf{99.4} & 88.0 & \textbf{98.4} & 93.6 & 91.1 & 52.5 &
47.4 & 79.1 & 85.4\gainu{$\uparrow$} \\

\rowcolor{black!5}& - &
78.8 & 86.8 & 93.6 & 94.2 & 91.5 & 95.3 &
98.8 & 76.2 & 97.0 & 84.6 & 71.4 & \textbf{96.5} &
99.3 & 87.4 & 98.3 & 93.2 & 91.3 & 51.6 &
50.2 & 79.4 & 85.8\phantom{0} \\

\rowcolor{black!5}\multirow{-2}{*}{ISO-CTS} & \checkmark &
79.5 &\textbf{ 87.1} & \textbf{95.0} & 96.5 & \textbf{91.6} & \textbf{95.8} &
\textbf{99.0} & \textbf{79.8} & \textbf{97.7} & 84.5 & \textbf{73.7} &
96.1 & \textbf{99.4} & \textbf{88.6} & \textbf{98.4} &
\textbf{94.2} & 91.4 & \textbf{56.4} &
60.0 & \textbf{81.5} & \textbf{87.3}\gainu{$\uparrow$} \\

& - &
64.4 & 56.6 & 49.1 & 42.1 & 67.2 & 56.8 &
95.3 & 46.2 & 64.9 & 78.3 & 54.9 & 91.3 &
95.7 & 62.9 & 92.9 & 70.5 & 82.0 & 19.9 &
10.9 & 56.9 & 63.0\phantom{0} \\

\multirow{-2}{*}{TIES} & \checkmark &
67.2 & 61.8 & 68.2 & 67.8 & 82.4 & 71.7 &
98.0 & 54.4 & 79.7 & 73.9 & 61.1 & 92.3 &
98.3 & 75.1 & 96.6 & 81.1 & 86.0 & 29.7 &
11.8 & 51.7 & 70.4\gainu{$\uparrow$} \\

\rowcolor{black!5}& - &
74.4 & 82.0 & 87.8 & 89.9 & 90.5 & 90.6 &
98.6 & 65.5 & 92.8 & \textbf{86.1} & 67.7 & 96.0 &
98.9 & 83.0 & 97.8 & 91.2 & 91.4 & 45.7 &
56.2 & 77.9 & 83.2\phantom{0} \\

\rowcolor{black!5}\multirow{-2}{*}{TSV-M} & \checkmark &
76.5 & 84.8 & 91.6 & \textbf{96.8} &
90.1 & 92.5 &
98.8 & 73.5 & 96.3 & 84.6 & 71.7 & 96.2 &
99.2 & 86.1 & 98.3 & 93.6 & \textbf{91.9} &
53.8 & \textbf{69.8} & 81.4 & 86.4\gainu{$\uparrow$} \\

\bottomrule
\end{tabular}}
\caption{
\textbf{Performance on 20 classification tasks for each ViT backbone.}
We evaluate five merging methods, with and without LARV, across 20 diverse datasets spanning natural, synthetic, texture, medical, and fine-grained categories. 
Task IDs correspond to: 
\#1---SUN397, \#2---Stanford Cars, \#3---RESISC45, \#4---EuroSAT, \#5---SVHN, 
\#6---GTSRB, \#7---MNIST, \#8---DTD, \#9---Flowers102, \#10---PCAM, 
\#11---FER2013, \#12---Pets, \#13---STL10, \#14---CIFAR100, \#15---CIFAR10, 
\#16---Food101, \#17---Fashion-MNIST, \#18---EMNIST-Letters, \#19---KMNIST, 
\#20---Rendered-SST2. 
Fine-tuned models serve as an upper bound. 
Best results per task (excluding Fine-tuned) are highlighted in bold.
These detailed results extend the aggregated findings in the main text by showing that LARV provides consistent gains across a significantly broader task distribution.
}

\end{table*}


\end{document}